\documentclass[conference,letterpaper]{IEEEtran}
\IEEEoverridecommandlockouts
\usepackage{cite}
\usepackage{amsmath,amssymb,amsfonts}
\usepackage{algorithmic}
\usepackage{algorithm}
\usepackage{graphicx}
\usepackage{textcomp}
\usepackage{xcolor}
\usepackage{array}
\usepackage[caption=false,font=normalsize,labelfont=sf,textfont=sf]{subfig}
\usepackage{url}
\usepackage{tikz}
\usetikzlibrary{arrows.meta, positioning, shadows.blur}

\def\BibTeX{{\rm B\kern-.05em{\sc i\kern-.025em b}\kern-.08em
    T\kern-.1667em\lower.7ex\hbox{E}\kern-.125emX}}

\begin{document}

\title{RRNet: Configurable Real-Time Video Enhancement with Arbitrary Local Lighting Variations}

\author{
\IEEEauthorblockN{%
Wenlong Yang\IEEEauthorrefmark{1}\thanks{The author is a Microsoft employee. This paper results from the author's academic research pursuing his PhD degree at Tsinghua University.}\,
Canran Jin,
Weihang Yuan,
Chao Wang,
Lifeng Sun\IEEEauthorrefmark{1}}
\IEEEauthorblockA{\IEEEauthorrefmark{1}Tsinghua University, Beijing, China\\
\texttt{ywl22@mails.tsinghua.edu.cn}, \texttt{sunlf@tsinghua.edu.cn}}
}

\maketitle

\begin{abstract}
With the growing demand for real-time video enhancement in live applications, existing methods often struggle to balance speed and effective exposure control, particularly under uneven lighting. We introduce RRNet (Rendering Relighting Network), a lightweight and configurable framework that achieves a state-of-the-art tradeoff between visual quality and efficiency. By estimating parameters for a minimal set of virtual light sources, RRNet enables localized relighting through a depth-aware rendering module without requiring pixel-aligned training data. This object-aware formulation preserves facial identity and supports real-time, high-resolution performance using a streamlined encoder and lightweight prediction head. To facilitate training, we propose a generative AI-based dataset creation pipeline that synthesizes diverse lighting conditions at low cost. With its interpretable lighting control and efficient architecture, RRNet is well suited for practical applications such as video conferencing, AR-based portrait enhancement, and mobile photography. Experiments show that RRNet consistently achieves the best overall result in low-light enhancement, localized illumination adjustment, and glare removal.
\end{abstract}


\section{Introduction}
The growing popularity of live streaming, video conferencing, and virtual reality has increased the demand for real-time video enhancement under low-light or uneven illumination. In such scenarios, suboptimal lighting (e.g., backlight or localized glare) often leads to underexposed videos with reduced visibility, loss of detail, and inconsistent appearance, degrading both visual quality and communication effectiveness. Effective exposure correction thus requires simultaneously achieving high visual quality, real-time performance, and computational efficiency, which remains challenging for high-resolution and mobile deployment.

\begin{figure}[!t]
    \centering
    \includegraphics[width=0.48\textwidth]{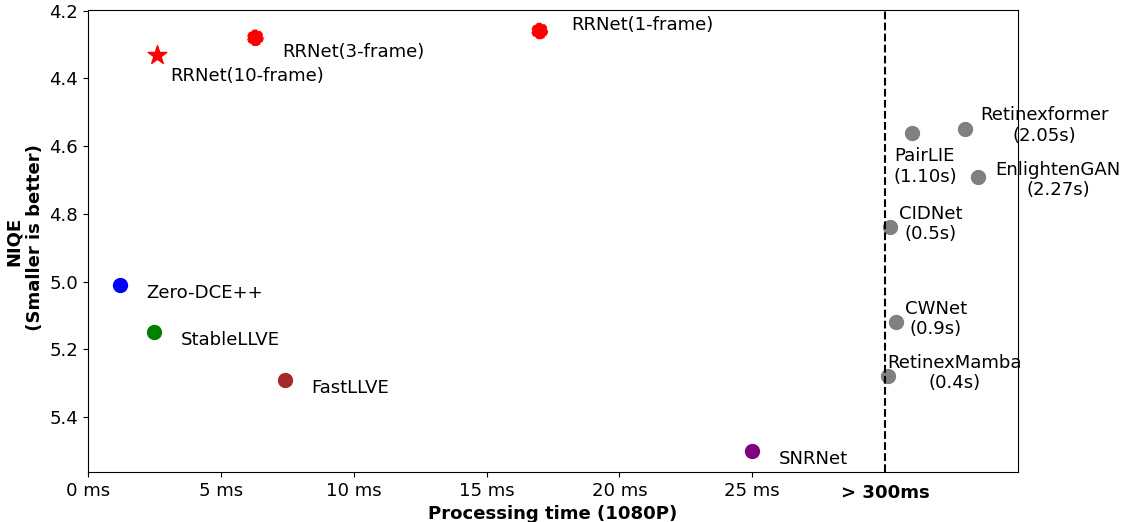}
    \caption{NIQE--runtime comparison on 1080p VCD videos using an RTX 3090 (lower is better). RRNet (\(N\)-frame) estimates lighting every \(N\) frames and reuses it between updates, amortizing the estimation cost. For \(N=1,3,10\), runtime is 17.0, 6.3, and 2.6\,ms per frame, respectively.}
    \label{fig:processing_time_NIQE}
\end{figure}

Figure~\ref{fig:processing_time_NIQE} summarizes the visual quality and efficiency of our method compared to state-of-the-art approaches on 1080p input video, as discussed later in the paper.

\begin{figure*}[t]
    \centering
    \setlength{\tabcolsep}{1pt}
    \begin{tabular}{cccccc}
        \includegraphics[width=0.14\textwidth]{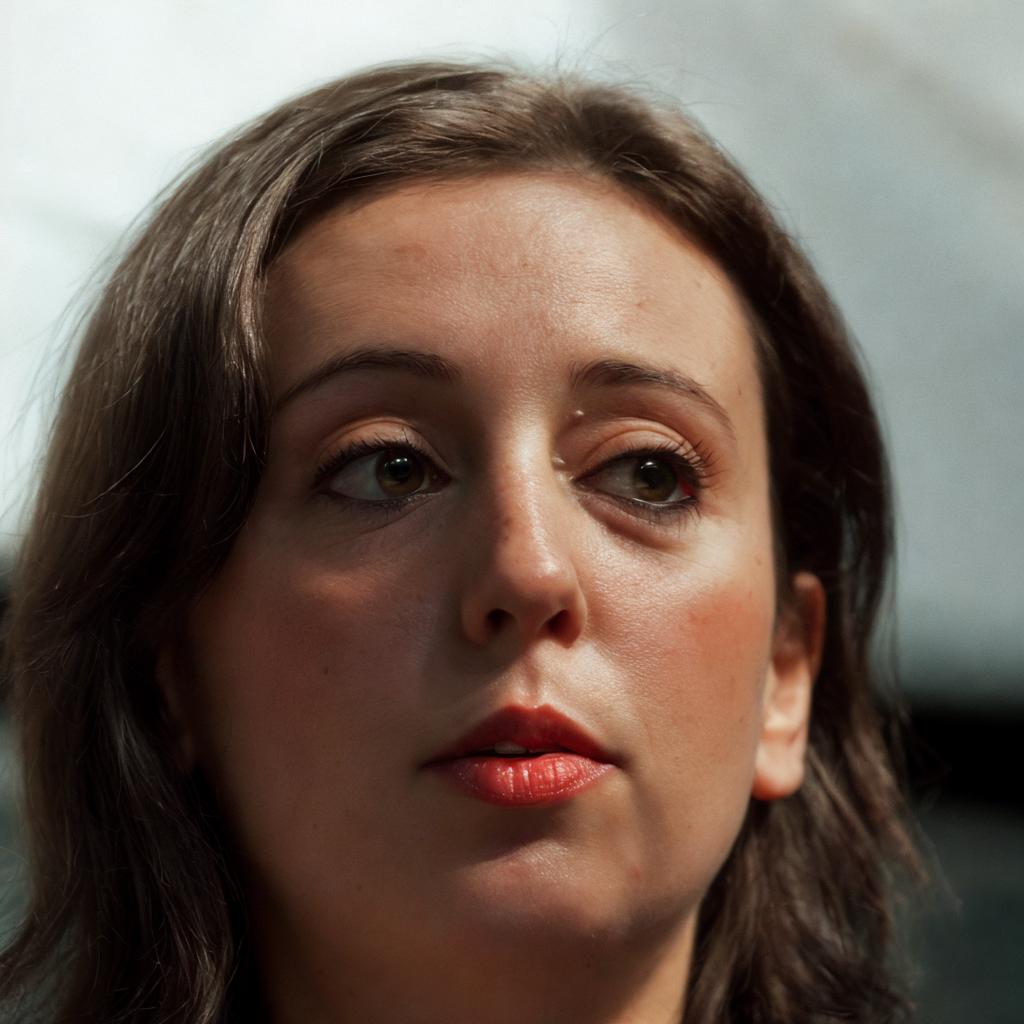} &
        \includegraphics[width=0.14\textwidth]{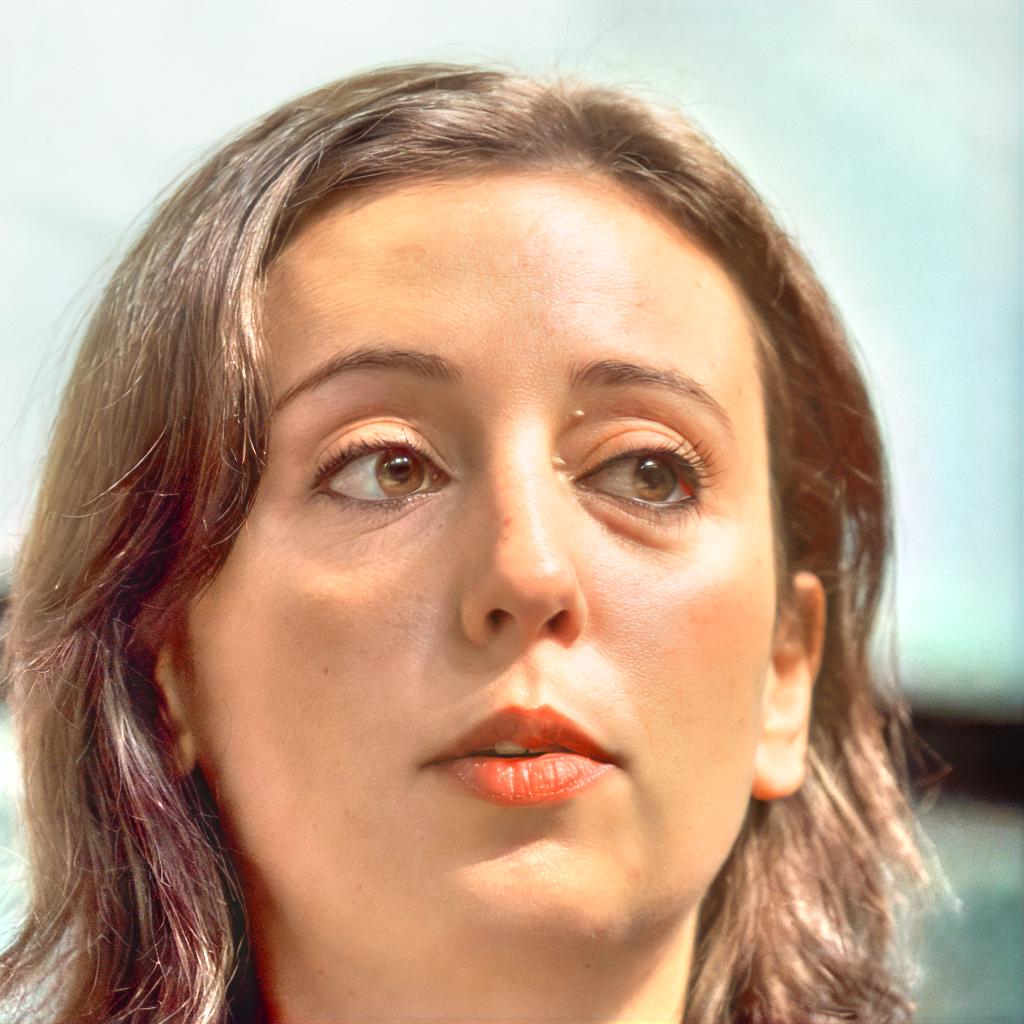} &
        \includegraphics[width=0.14\textwidth]{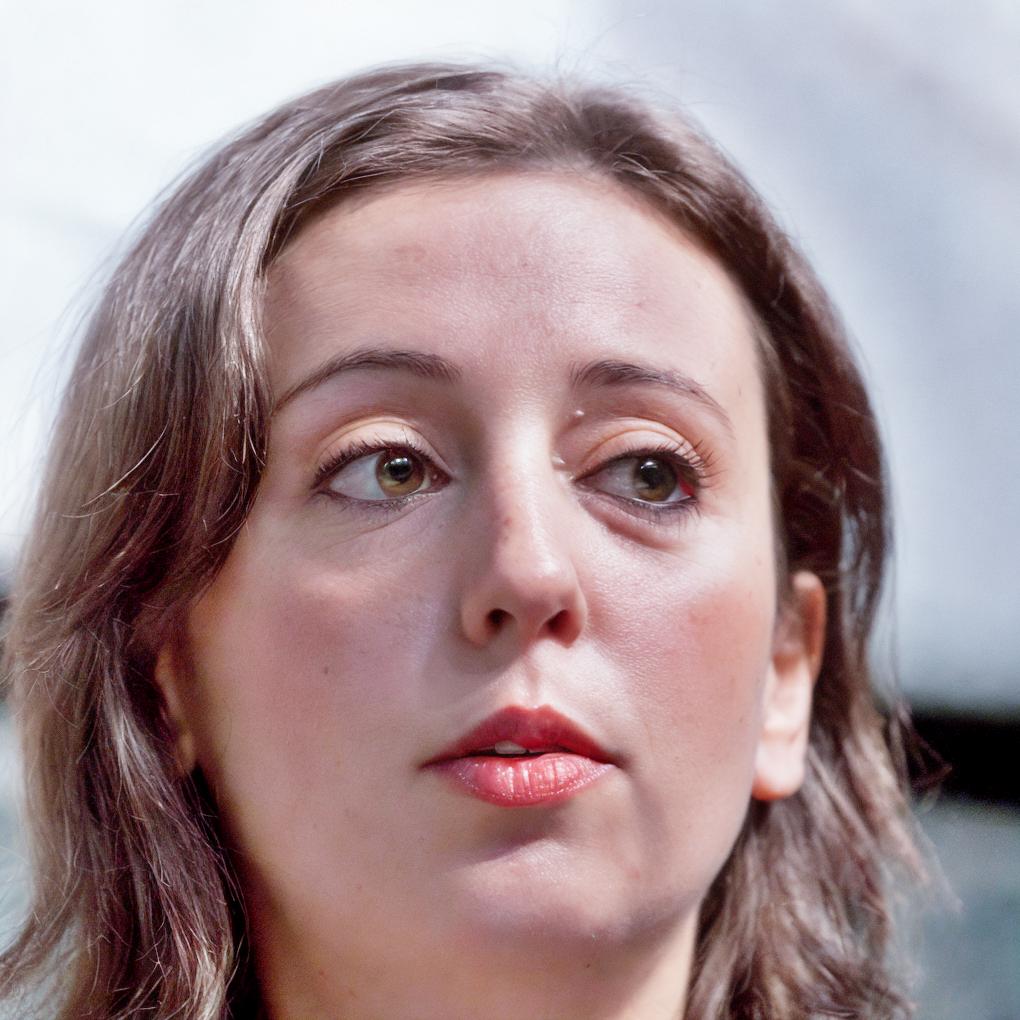} &
        \includegraphics[width=0.14\textwidth]{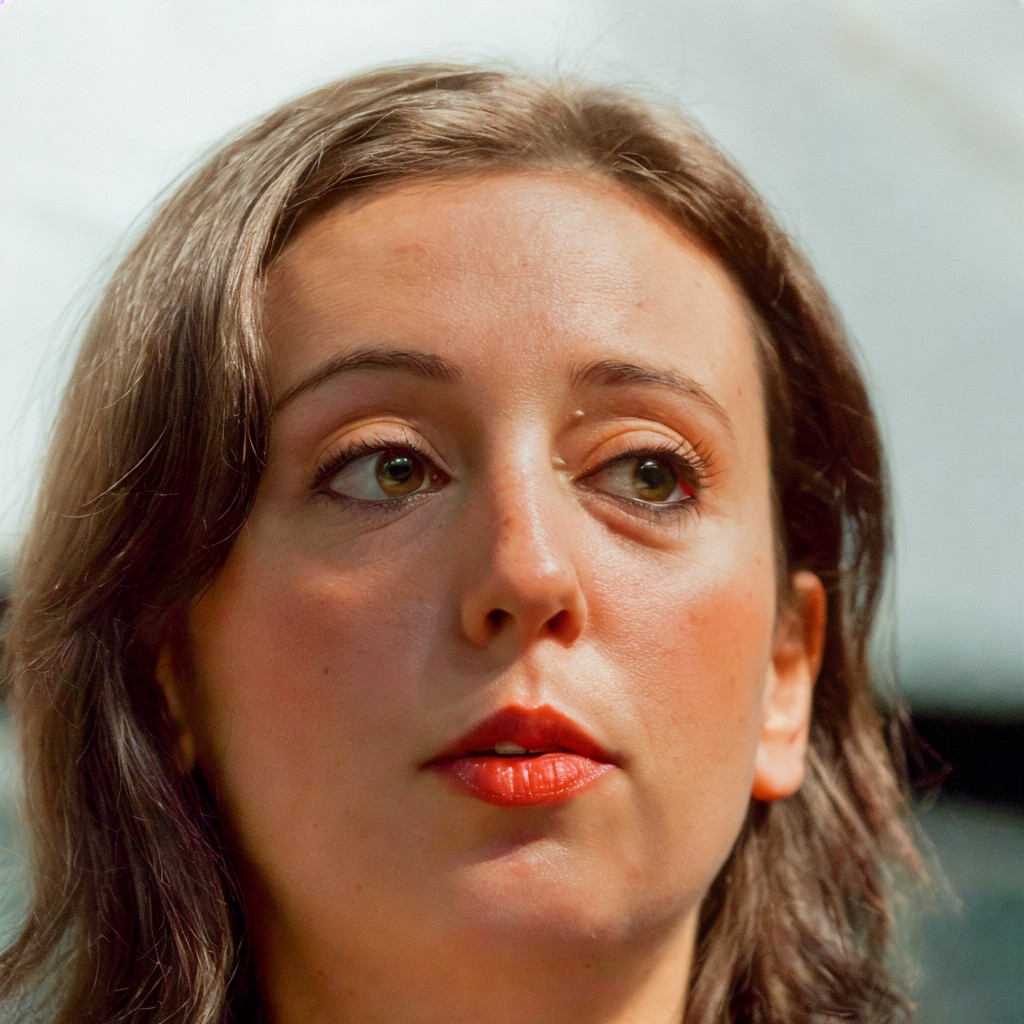} &
        \includegraphics[width=0.14\textwidth]{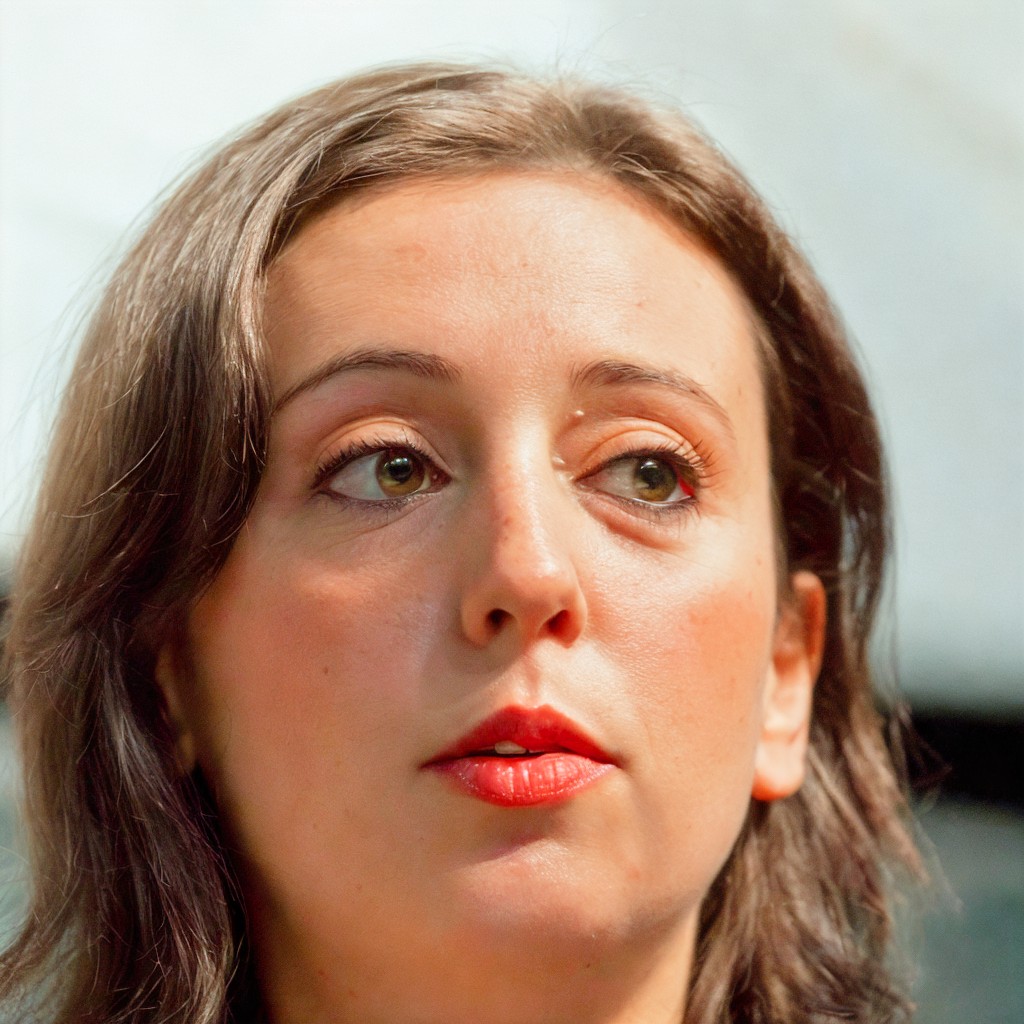} &
        \includegraphics[width=0.14\textwidth]{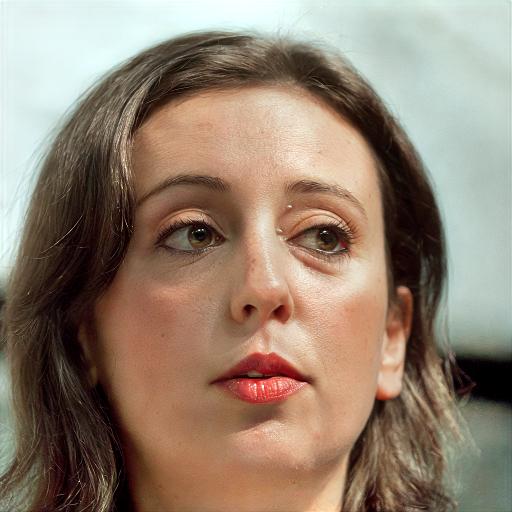} \\
        \includegraphics[width=0.14\textwidth]{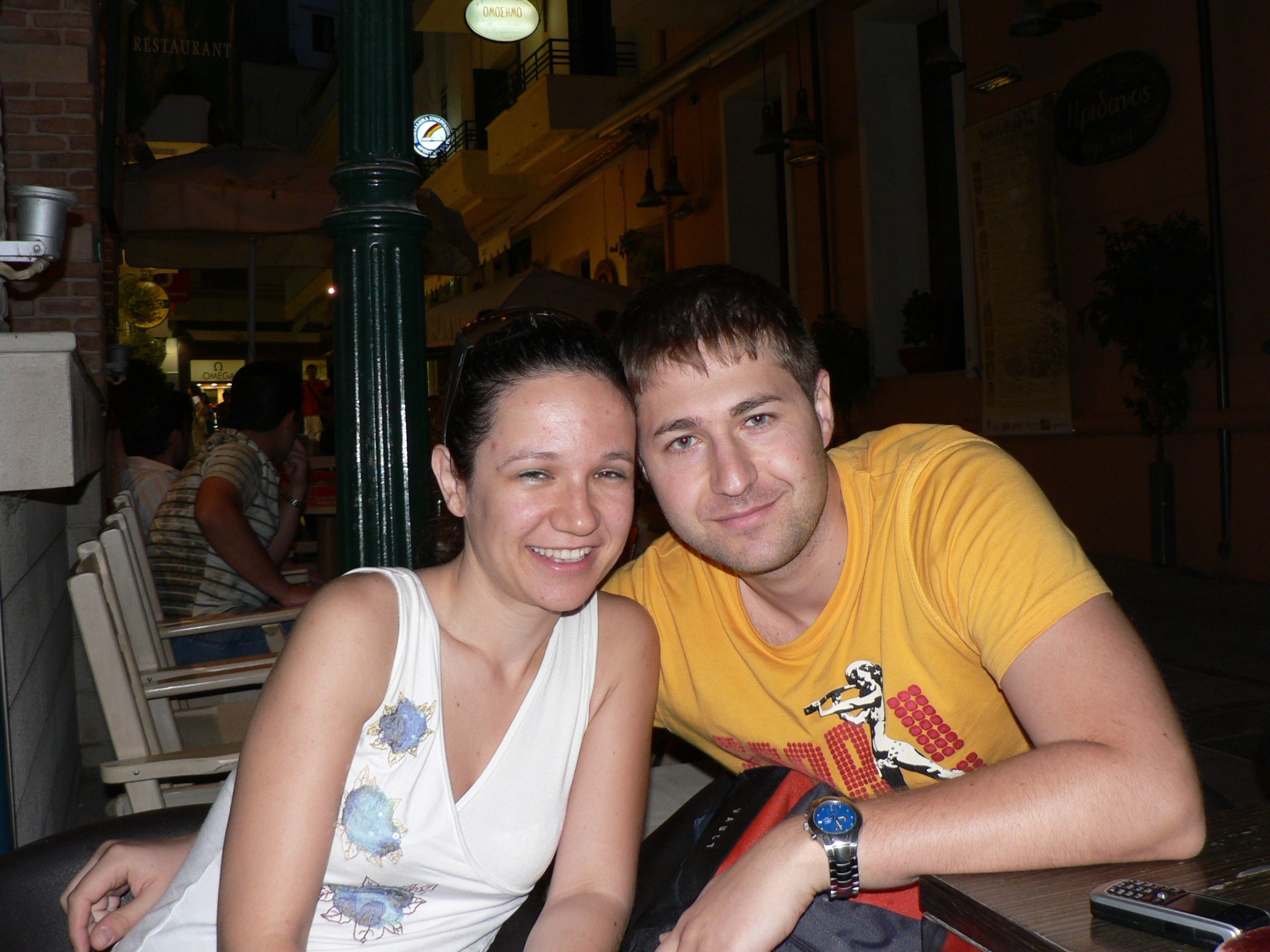} &
        \includegraphics[width=0.14\textwidth]{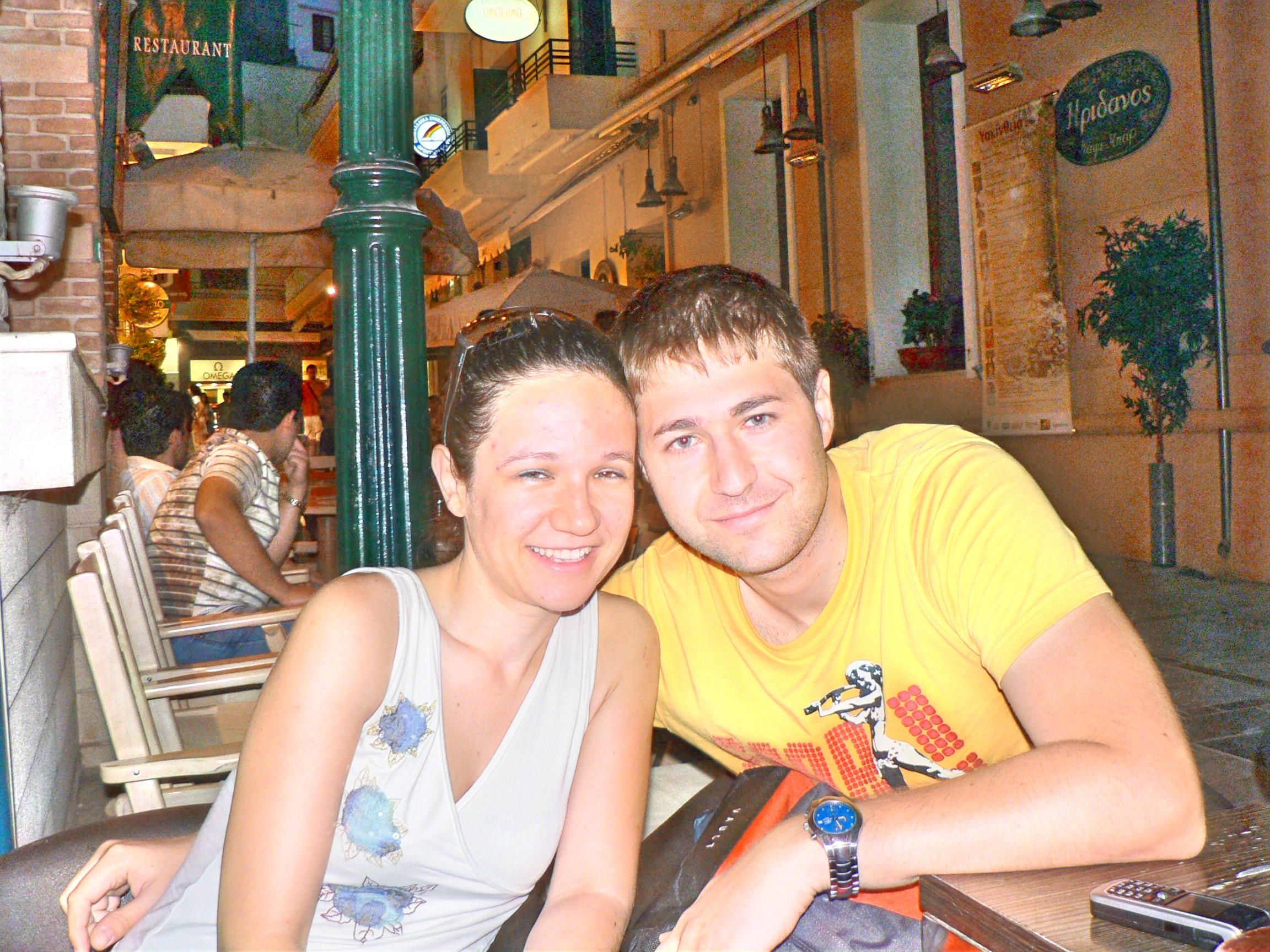} &
        \includegraphics[width=0.14\textwidth]{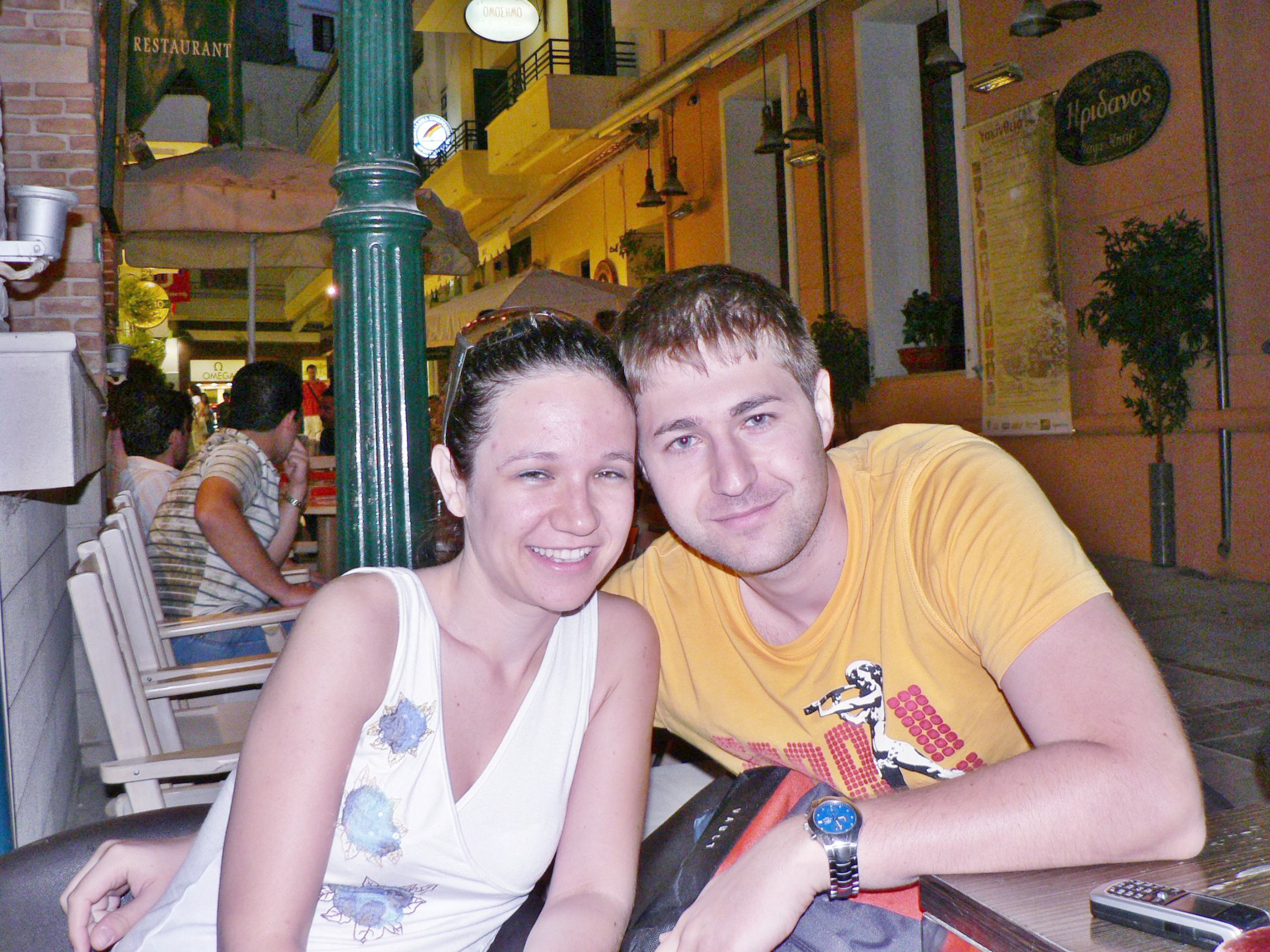} &
        \includegraphics[width=0.14\textwidth]{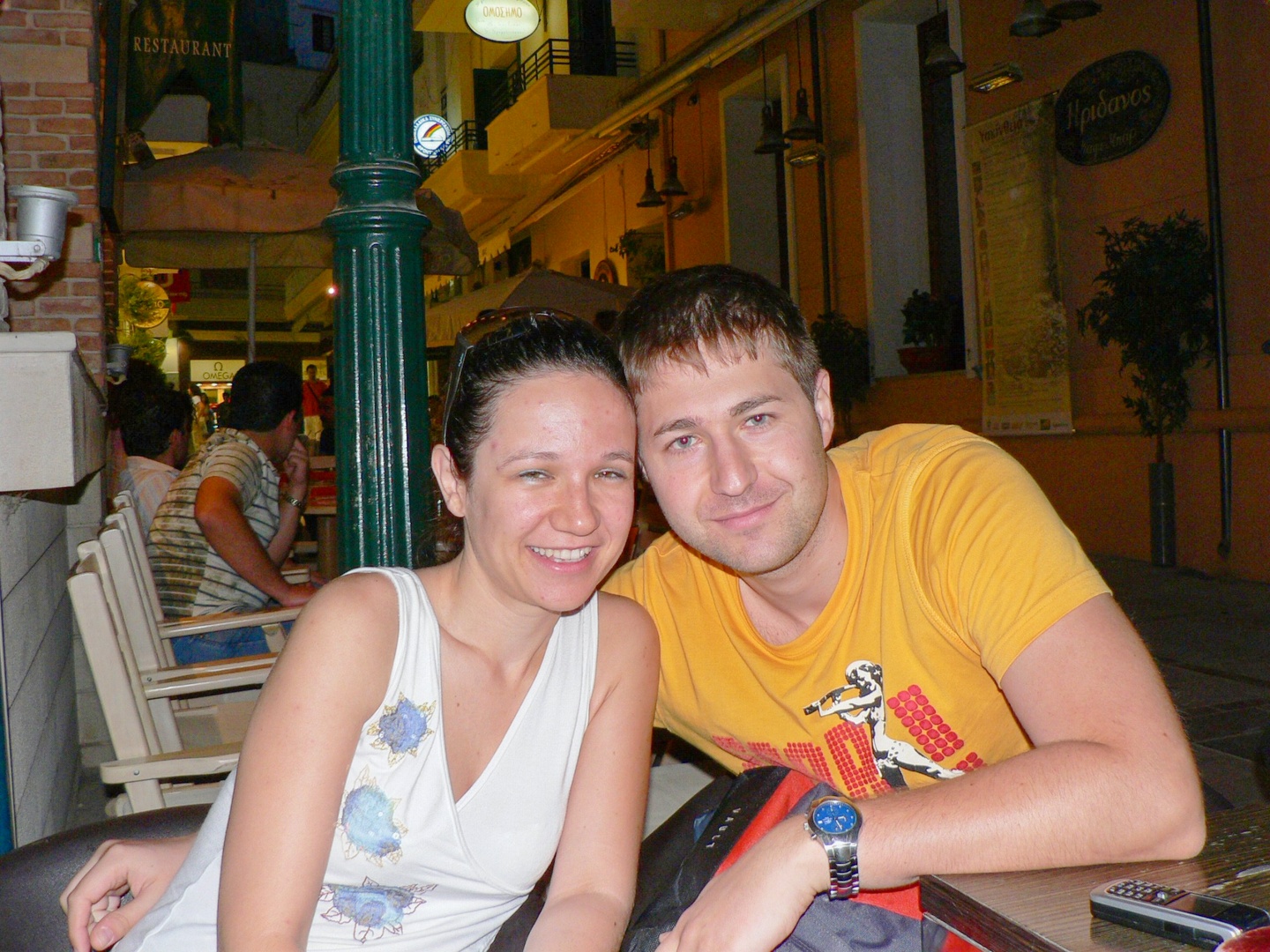} &
        \includegraphics[width=0.14\textwidth]{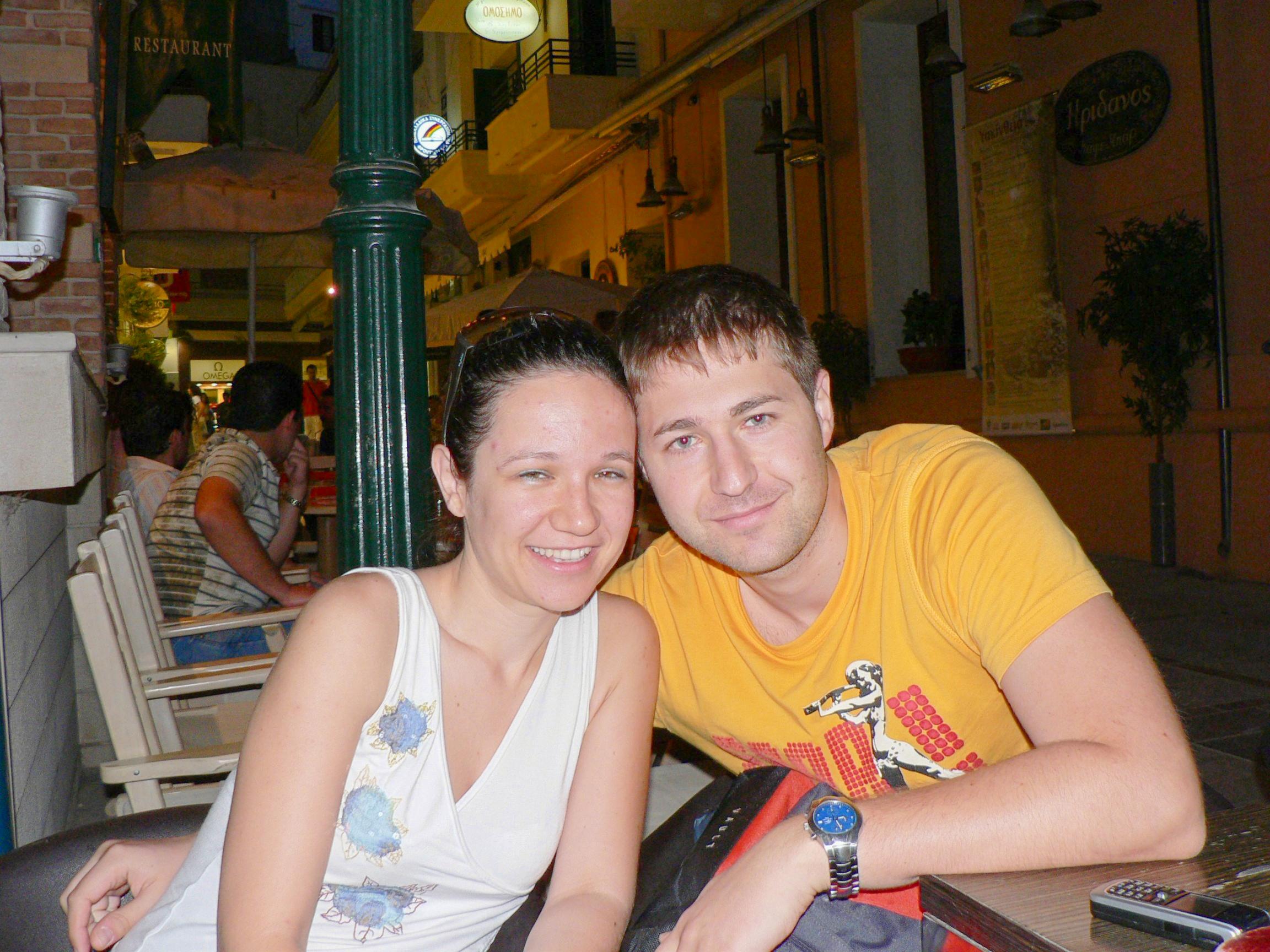} &
        \includegraphics[width=0.14\textwidth]{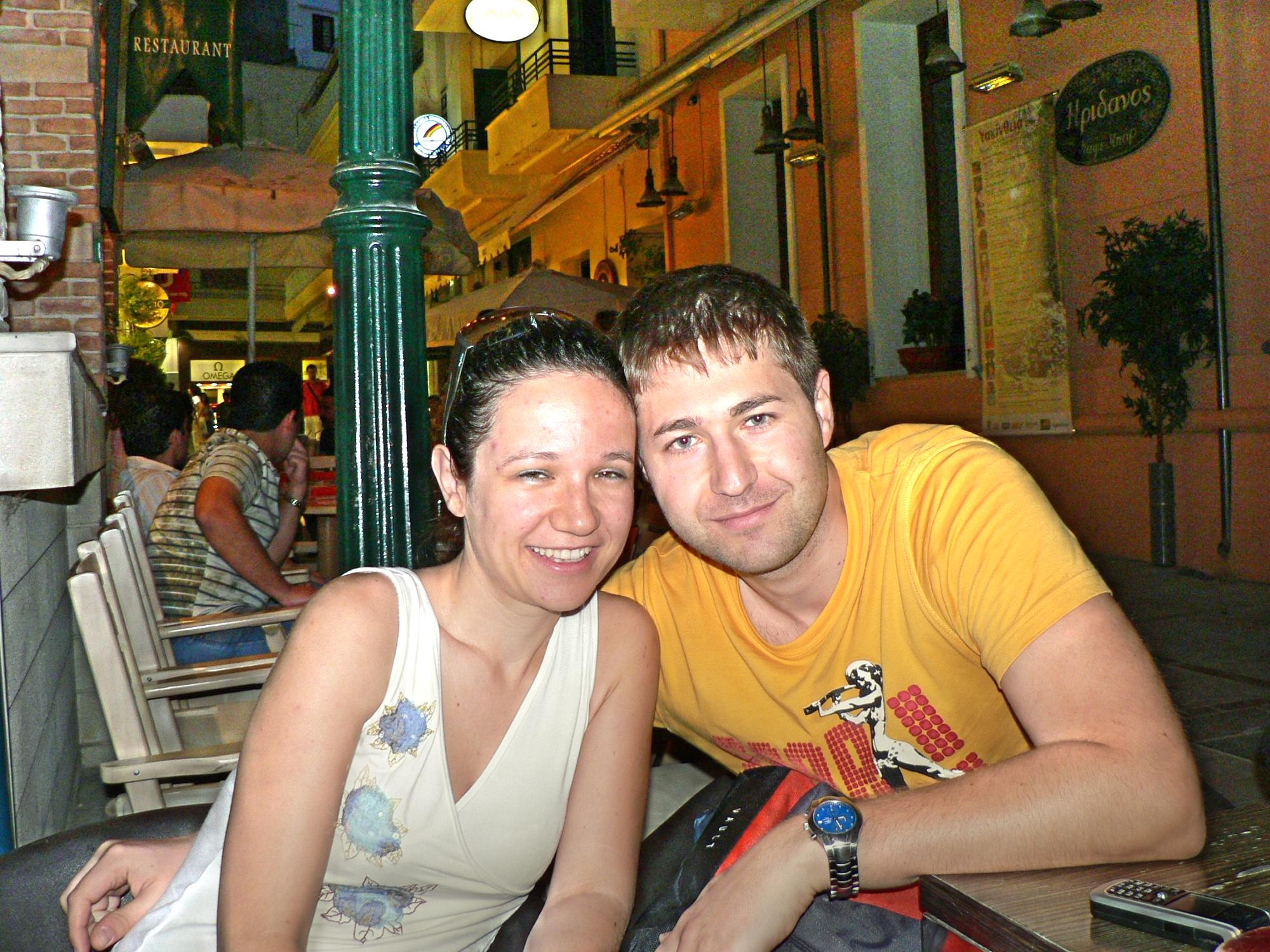} \\
        \includegraphics[width=0.14\textwidth]{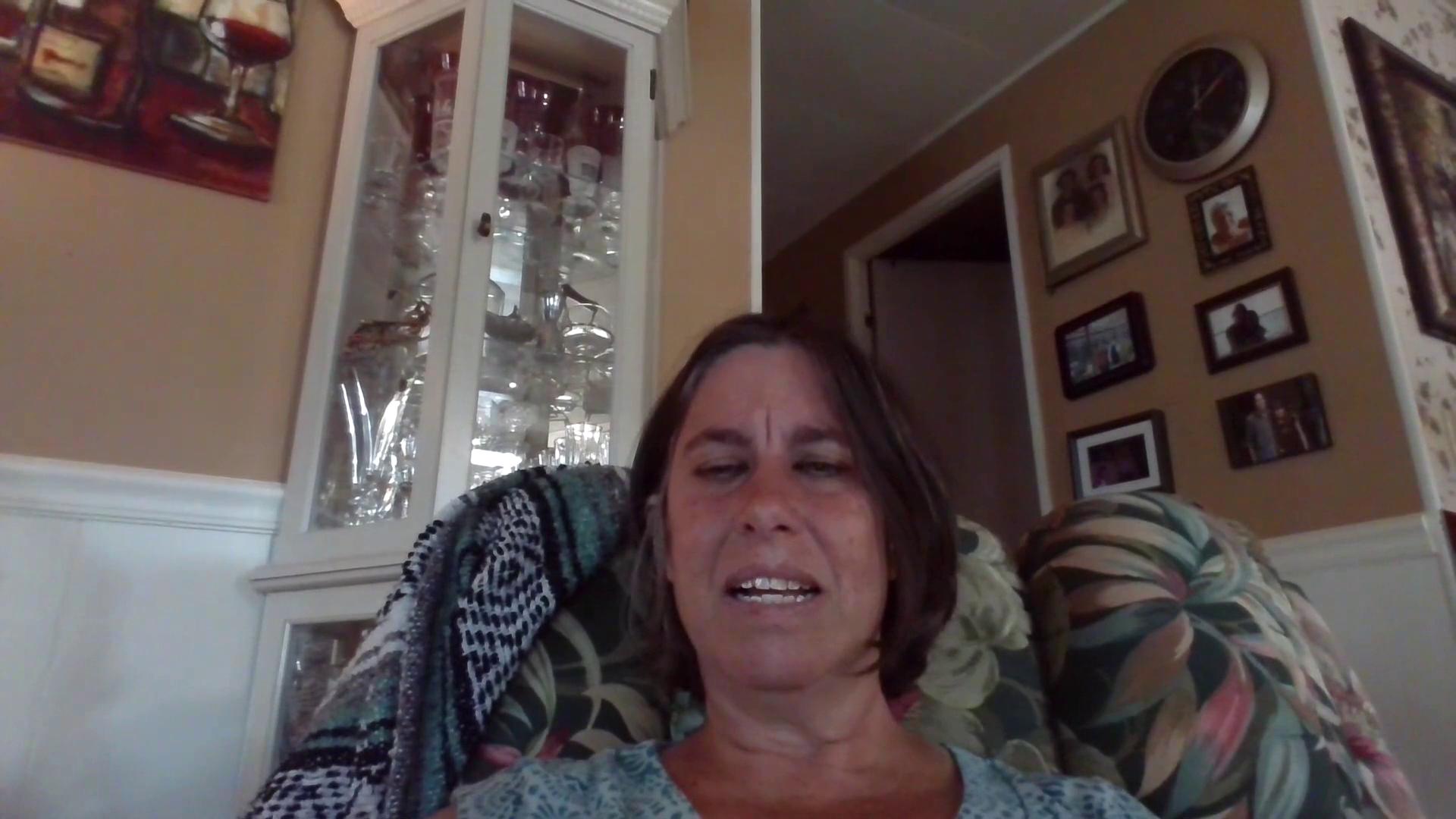} &
        \includegraphics[width=0.14\textwidth]{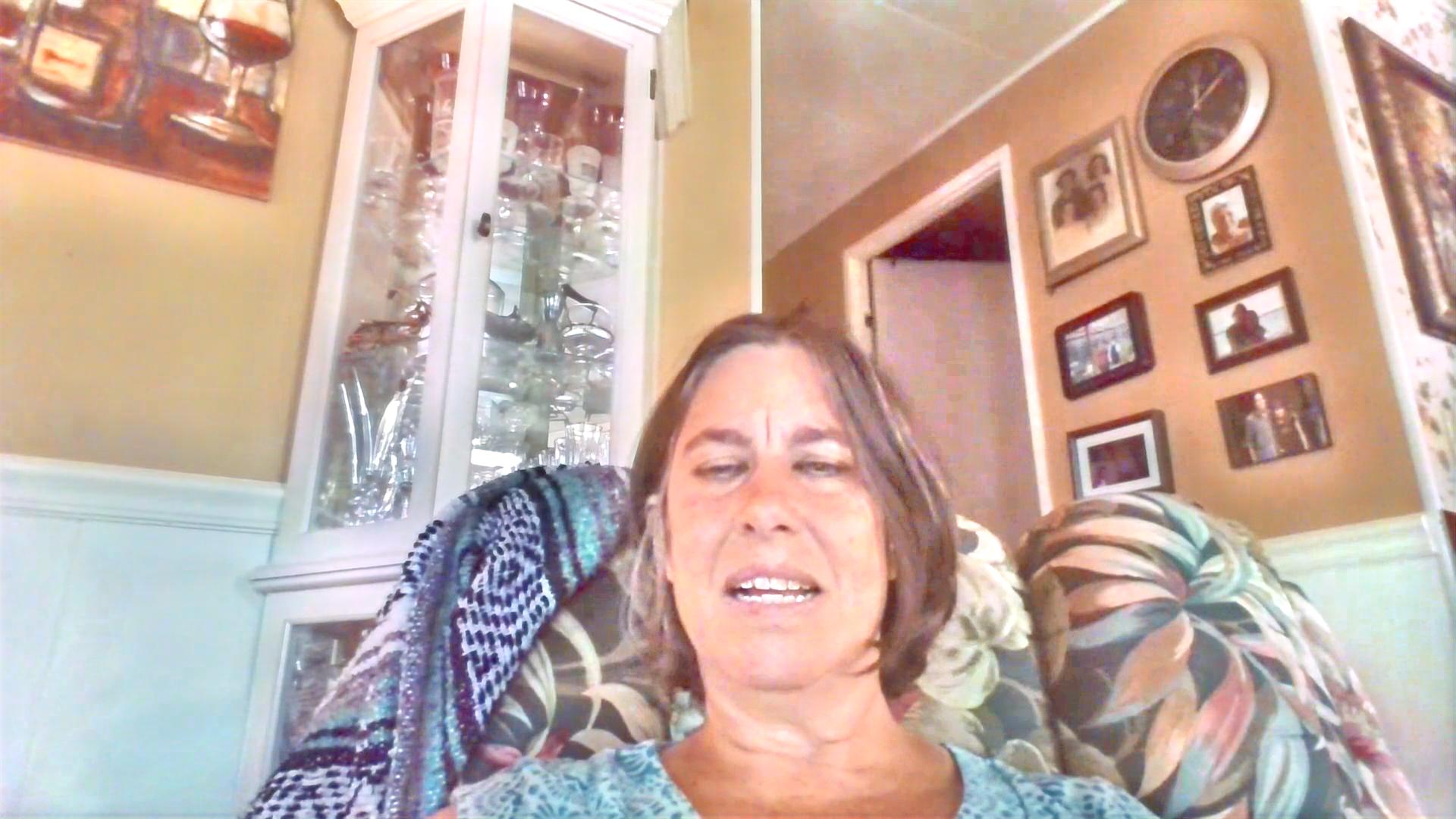} &
        \includegraphics[width=0.14\textwidth]{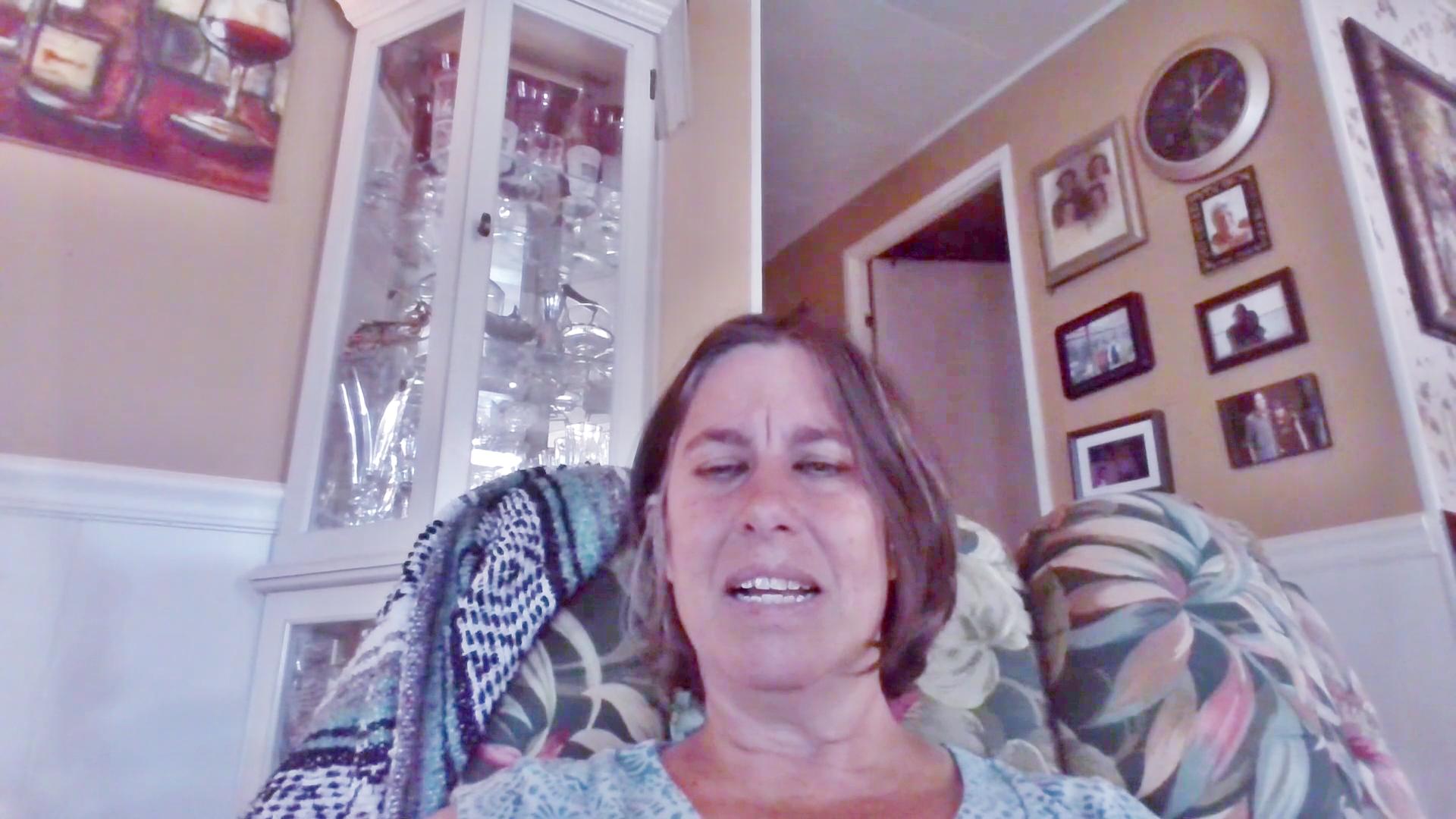} &
        \includegraphics[width=0.14\textwidth]{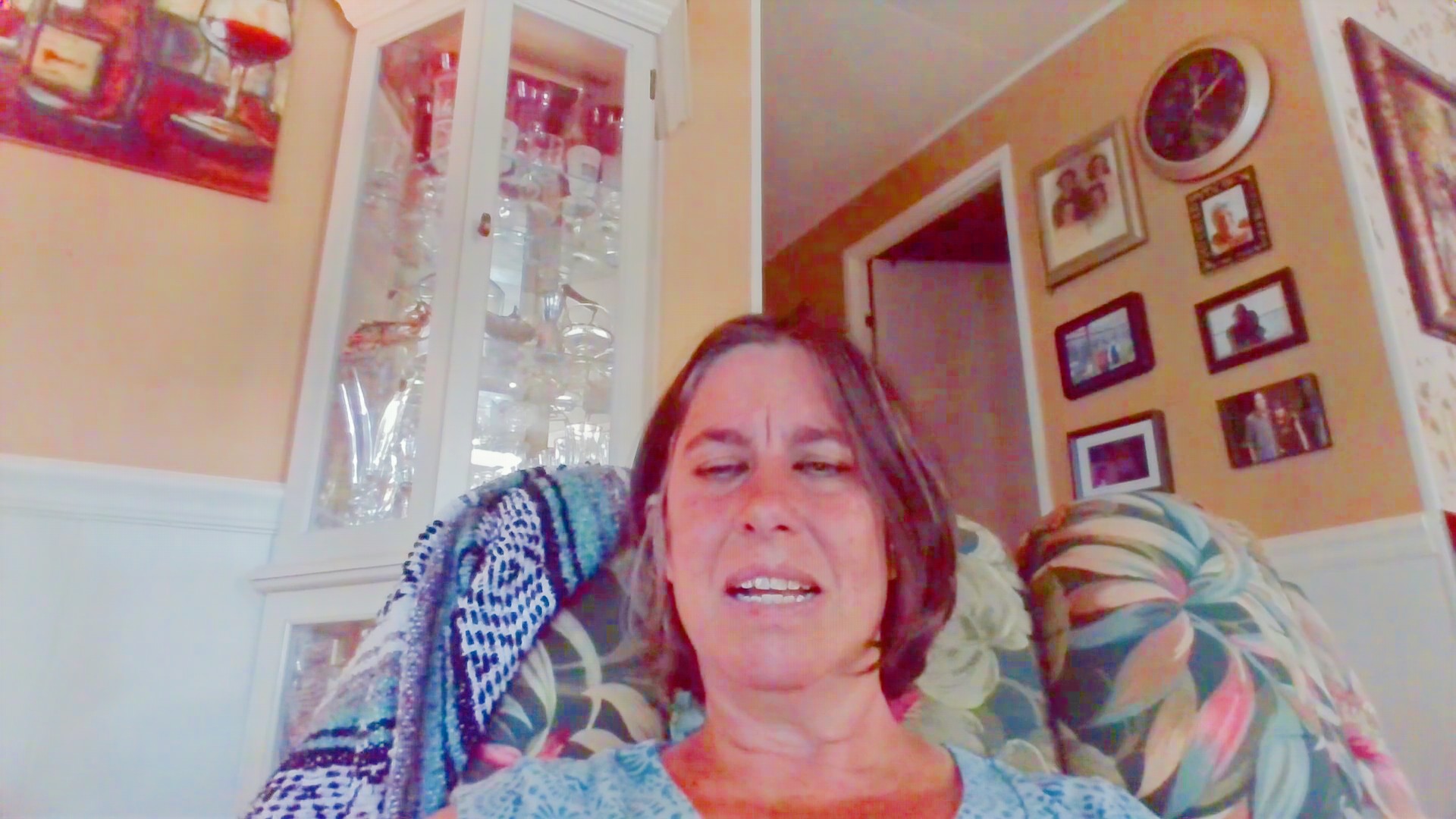} &
        \includegraphics[width=0.14\textwidth]{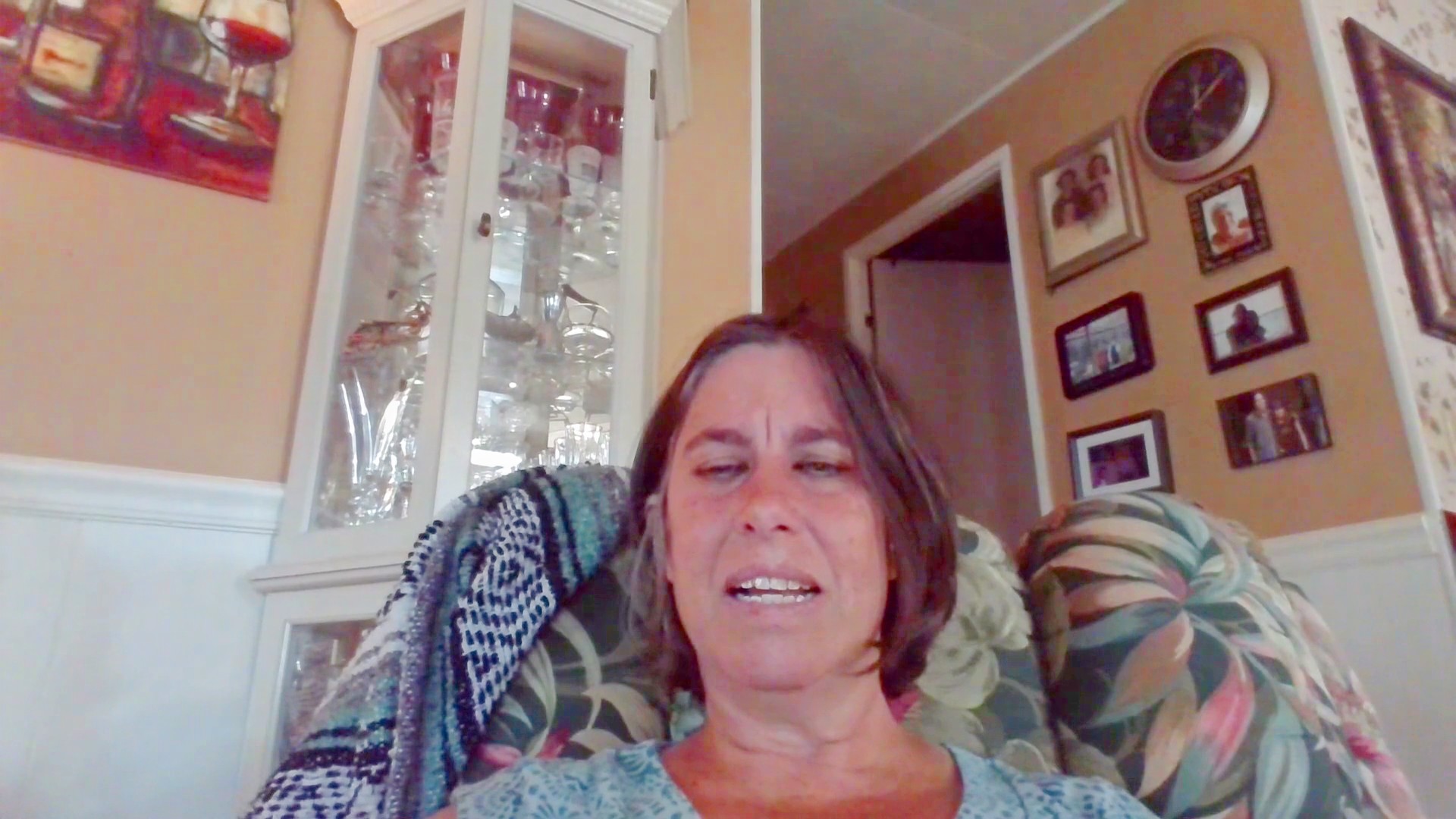} &
        \includegraphics[width=0.14\textwidth]{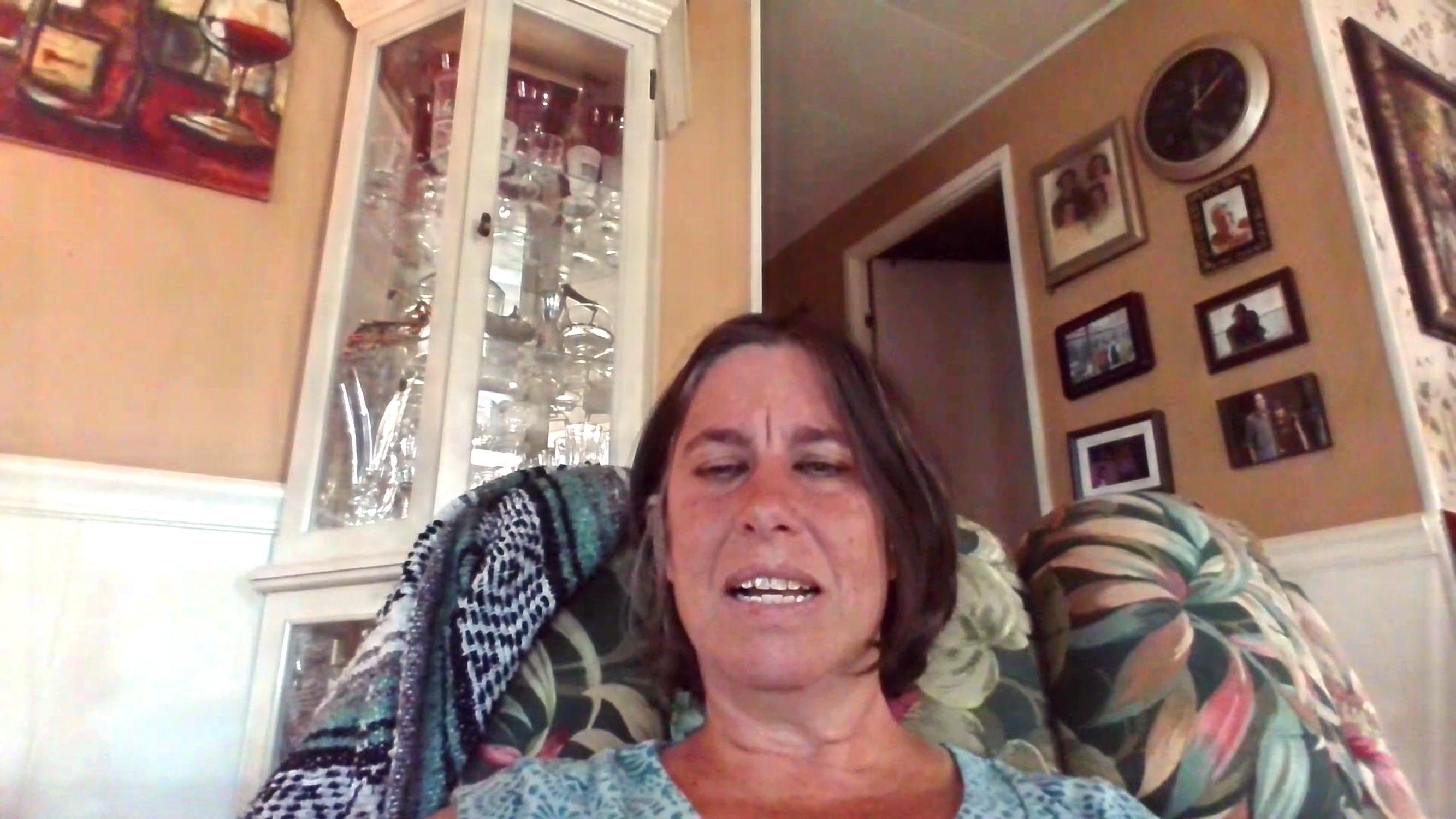} \\
        Input & EnlightenGAN & Zero-DCE++ & CWNet & CIDNet & RRNet (Ours) \\
    \end{tabular}
    \caption{VisualComparisons of various low-light enhancement methods on test images from the FFHQ, VV, and VCD datasets show that RRNet achieves superior exposure balance and preserves skin tones more effectively than other state-of-the-art methods.}
    \label{fig:visual_comparison}
\end{figure*}

Classical global adjustments (histogram equalization, gamma correction~\cite{gamma2iccv}) often over-enhance or distort color, while Retinex-based models~\cite{LIME, retinexformer} are sensitive to their smoothness assumptions. Recent deep methods~\cite{retinexmambabai2024, cidnetyan2025hvi, cwnetzhang2025cwnet_iccv} improve quality but rely on dense pixel-wise prediction with high compute. Lightweight video models~\cite{fastllve, StableLLVE} hit real-time speed via global or grid-based adjustments yet lack object awareness and produce artifacts under complex lighting.

We propose \textbf{RRNet (Rendering Relighting Network)}, a lightweight framework for real-time video enhancement under complex illumination. Instead of synthesizing pixels, RRNet predicts virtual lighting parameters and renders depth-aware shading, removing heavy decoders. By estimating a minimal set of virtual light sources, it performs object-aware localized relighting with temporal coherence across frames, evaluated on image and video benchmarks via NIQE~\cite{mittal2012niqe}.

Our main contributions are summarized as follows:
\begin{itemize}
    \item A \textbf{lightweight real-time video enhancement framework} based on virtual lighting, with a physically motivated lighting parameter regularization to ensure stable and plausible relighting.
    \item A depth-aware rendering module for \textbf{object-aware lighting control} under uneven illumination, preserving facial identity in videos.
    \item A \textbf{generative AI-based dataset pipeline} for scalable local relighting training without pixel-level paired data.
\end{itemize}

\section{Related Work}

RRNet targets real-time video enhancement under arbitrary local lighting variations, which sits between efficient video enhancement / portrait relighting and low-light image enhancement (LLIE). Since no public benchmark provides paired ground truth for arbitrary local relighting on live video, we evaluate on widely-used LLIE benchmarks whose scenes also exhibit spatially uneven illumination, enabling broad SOTA comparison.

\textbf{Real-Time Video Enhancement and Relighting.}
Real-time video enhancement methods prioritize efficiency and temporal coherence. FastLLVE~\cite{fastllve} uses intensity-aware lookup tables, while StableLLVE~\cite{StableLLVE} incorporates temporal smoothing constraints. Although effective, these methods struggle with uneven lighting and fine-grained object boundaries.

\textbf{Portrait Relighting.}
Portrait relighting methods~\cite{kim2024switchlight} typically employ inverse rendering formulations to decompose shading and reflectance, followed by neural rendering or diffusion-based synthesis. While producing high-quality results, their computational cost — typically requiring multi-step diffusion sampling or dense per-pixel decoders — limits real-time applicability. In contrast, RRNet performs a single forward pass that regresses a compact set of virtual-light parameters, retaining interpretable lighting control at video frame rates.

\textbf{Low-Light Image Enhancement.}
Traditional methods, such as histogram equalization and gamma correction~\cite{gamma2iccv}, perform global intensity adjustment and often fail under spatially non-uniform illumination. Retinex-based approaches~\cite{Land71} decompose images into reflectance and illumination, and recent deep variants such as Retinexformer~\cite{retinexformer} and RetinexMamba~\cite{retinexmambabai2024} improve global consistency and efficiency, yet remain sensitive to the underlying smoothness assumptions when illumination is complex or multi-source. Learning-based methods further address LLIE by jointly correcting illumination and denoising; representative works include RetinexNet~\cite{Wei18}, PairLIE~\cite{PairLIE}, and SNRNet~\cite{snrnet}. Zero-DCE~\cite{Zero-DCE++} introduces a lightweight curve-estimation alternative without paired data but applies global adjustments, while EnlightenGAN~\cite{Jiang19} enables unsupervised enhancement but suffers from instability and is less suitable for video. More recent models such as CIDNet~\cite{cidnetyan2025hvi} and CWNet~\cite{cwnetzhang2025cwnet_iccv} further explore efficient network designs for LLIE. These methods provide a strong comparison set, although none directly model the localized, object-aware relighting that motivates RRNet.

\section{Approach}
\label{sec:approach}

\begin{figure*}[htbp]
    \centering
    \includegraphics[width=0.9\textwidth]{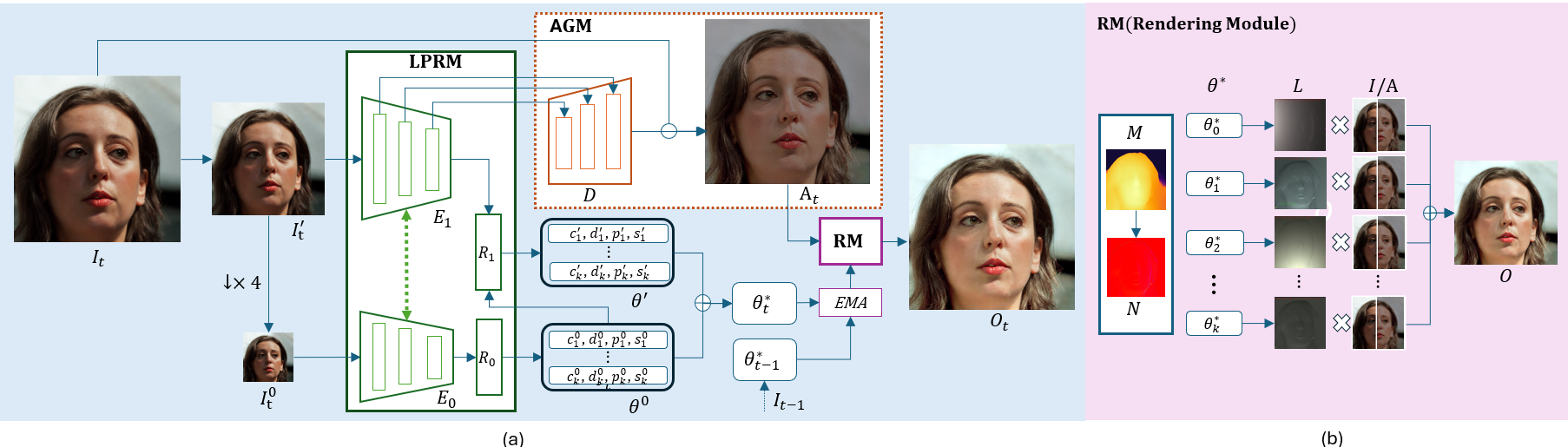}
    \caption{
        Architecture of \textbf{RRNet}.
        (a) Overall framework with \textit{Lighting Parameter Regression Module (LPRM)}, \textit{Rendering Module (RM)}, and optional \textit{Albedo Generation Module (AGM)}; the dashed line in \textit{LPRM} indicates shared weights between encoders $E_0$ and $E_1$.
        (b) \textit{RM} details; \(M\), \(N\), and \( \theta^* \) denote depth, surface normal, and estimated lighting parameters.
    }
    \label{fig:model_architecture}
\end{figure*}

RRNet performs real-time enhancement under complex illumination by predicting virtual lighting parameters and adjusting illumination via depth-aware rendering.

\subsection{Problem Formulation}
We reformulate pixel-level lighting enhancement as virtual lighting parameter regression. A simplified parallel-light model is defined as
\begin{equation}
\theta = \big\{ \{ c, d, p, s \}_k \big\}_{k=1}^{K} \ \cup \ \{ L_{\text{ambient}} \},
\label{eq:theta}
\end{equation}
where \( c_k \in \mathbb{R}^3 \) denotes color-wise intensity, \( d_k \) the light direction, \( p_k \) the light center in normalized coordinates, \( s_k \) the distance based attenuation factor, \( L_{\text{ambient}} \) the ambient term, and \(k \) the number of virtual lights.

\subsection{Architecture}
As shown in Fig.~\ref{fig:model_architecture}, RRNet consists of LPRM (lighting parameter regression), RM (rendering), and an optional AGM (albedo generation) for challenging static cases (e.g., glare). Unlike encoder--decoder image translation, RRNet predicts lighting parameters instead of pixels, removing heavy decoders and enabling real-time high-resolution processing.

\textbf{Lighting Parameter Regression (LPRM).}
Given $I$ resized to short side 512, a coarse branch predicts $\theta^0$ from a $4\times$ downsampled global view and a refined branch predicts an offset $\theta'$ from full-resolution features for local correction. Both branches share a RepViT~\cite{RepViT-SAM} encoder; $R_0$ reuses its classifier head, and $R_1$ concatenates $\theta^0$ with the pooled embedding to regress $\theta'$. We apply per-dimension reverse normalization with mean $\hat{\mu}$ and variance $\hat{\sigma}$ estimated from optimal parameters:
\begin{equation}
\theta^* = \hat{\sigma} \odot (\theta^0 + \theta') + \hat{\mu}.
\end{equation}

\textbf{Rendering Module (RM).}
We use a lightweight renderer based on a simplified Blinn--Phong model~\cite{Blinn1977}. For pixel \( i \),
\begin{equation}
L_k(i) = c_k \cdot \frac{\max(\sigma_2, \, N(i) \cdot d_k)}{s_k \| p_k - p_i \|^2 + \sigma_1},
\end{equation}
\begin{equation}
\begin{aligned}
L(i) &= L_{\text{ambient}} + \sum_{k=1}^{K} L_k(i), \\
O(i) &= I(i) \cdot L(i) \ \text{if}\ L(i) > \epsilon, \ \text{else}\ I(i),
\end{aligned}
\label{eq:render}
\end{equation}
where \( p_i=\{x, y, M(i)\} \) uses spatial coordinates and per-pixel depth \( M \); \( N \) is the normal derived from \( M \); \(\sigma_1,\sigma_2\) are stability constants; and the fallback branch ($\epsilon=0.05$) prevents degenerate dark outputs on out-of-distribution inputs without any additional training.

\textbf{Albedo Generation Module (AGM).}
For static images under strong directional lighting (e.g., glare), we optionally replace $I$ in RM with an estimated albedo $A = I - Z$, where the illumination mask $Z$ is upsampled from a 3-channel output $Z'$ of a lightweight U-Net decoder. Unless noted, RRNet uses the dual-branch LPRM with AGM and $K=9$ virtual lights.

\subsection{Temporal Smoothing Module}
\label{subsec:temporal_smoothing}
To improve temporal coherence, we smooth lighting parameters across frames using an exponential moving average:
\begin{equation}
    \theta_t^{\text{smooth}} = \beta \theta_{t-1}^{\text{smooth}} + (1 - \beta) \theta_t,
\label{eq:smoother}
\end{equation}
with \( \beta \in [0.8, 0.99] \). The smoothed parameters are then used in rendering to reduce flicker.

\subsection{Loss Functions}
\label{subsec:lossfunc}

RRNet is trained with
\begin{equation}
\mathcal{L}_{\text{total}} = \mathcal{L}_{\text{pixel}} + \mathcal{L}_{\text{roi}} + \lambda_r \mathcal{L}_{\text{reg}} + \lambda_{\text{ret}} \mathcal{L}_{\text{retinex}} ,
\end{equation}
combining four terms. The pixel loss is the L2 residual $\mathcal{L}_{\text{pixel}} = \| I_{\text{output}} - I_{\text{target}} \|_2$. The ROI loss reweights this residual by per-pixel weights derived from the depth map $M$ and the target luminance $I_{\text{target}}$, each clamped at $\sigma_c, \sigma_l$ so background pixels are not ignored:
\begin{equation}
\mathcal{L}_{\text{roi}} = \left\| \left[ \max(M, \sigma_c) \circ \max(I_{\text{target}}, \sigma_l) \right] \circ (I_{\text{output}} - I_{\text{target}}) \right\|_2^2 .
\end{equation}
The lighting regularization softly enforces valid parameter ranges:
\begin{equation}
\mathcal{L}_{\text{reg}} = \| \theta^{-} - \mathcal{C}(\theta^{-}) \|_2^2 + \lambda_{\text{amb}} \| L_{\text{ambient}} - \mathcal{C}_{\text{amb}}(L_{\text{ambient}}) \|_2^2 ,
\label{eq:lighting_reg_loss}
\end{equation}
where $\theta^{-}$ excludes the ambient term and $\mathcal{C}$ clamps positions/intensities and unit-normalizes directions. Finally, the Retinex consistency loss aligns the predicted reflectance with the target under the Retinex framework $I = R \cdot L$. We estimate a low-frequency illumination $L(x) = \mathrm{smooth}_k(\max_c I_c(x))$ with a large mean filter ($k=51$) and recover $R(x) = I(x)/(L(x)+\epsilon_r)$ ($\epsilon_r=0.02$, clipped to $[0,8]$), giving
\begin{equation}
\mathcal{L}_{\text{retinex}} = \| R(I_{\text{output}}) - R(I_{\text{target}}) \|_1 ,
\end{equation}
which directly matches the relighting objective of modulating illumination while preserving reflectance. We set $\lambda_{\text{ret}} = 0.05$.

\section{Dataset and Metrics}
\label{sec:dataset}

\subsection{FFHQL Dataset}
\label{subsec:construct_dataset}

\begin{figure*}[!t]
    \centering
    \begin{tikzpicture}[
        node distance=3mm and 3mm,
        every node/.style={font=\scriptsize\sffamily},
        stage/.style={rectangle, line width=0.4pt, rounded corners=2pt,
                      align=center, inner xsep=5pt, inner ysep=3pt,
                      minimum height=7mm,
                      blur shadow={shadow blur steps=4, shadow xshift=0.3pt,
                                   shadow yshift=-0.3pt, shadow opacity=12}},
        sub/.style={font=\tiny\itshape\sffamily, text=black!70, align=left,
                    inner sep=1pt},
        arrow/.style={-{Latex[length=2mm]}, thick, draw=black!70}
    ]
        \node[stage, draw=black!50, fill=black!4]
            (s1) {FFHQ portrait};
        \node[stage, draw=blue!55!black, fill=blue!7, right=of s1]
            (s2) {Lighting sampler};
        \node[stage, draw=blue!70!black, fill=blue!12, right=of s2]
            (s3) {IC-Light diffusion};
        \node[stage, draw=blue!55!black, fill=blue!7, right=of s3]
            (s4) {$N$ variants};
        \node[stage, draw=black!50, fill=black!4, right=of s4]
            (s5) {Expert voting};
        \node[stage, draw=teal!55!black, fill=teal!10, right=of s5]
            (out) {Training pair};

        \draw[arrow] (s1) -- (s2);
        \draw[arrow] (s2) -- (s3);
        \draw[arrow] (s3) -- (s4);
        \draw[arrow] (s4) -- (s5);
        \draw[arrow] (s5) -- (out);

        \node[below=1mm of s2, inner sep=1pt, align=center] (sub2) {%
            \setlength{\tabcolsep}{1pt}\renewcommand{\arraystretch}{0.85}%
            \begin{tabular}{cccc}
                \includegraphics[width=5.5mm]{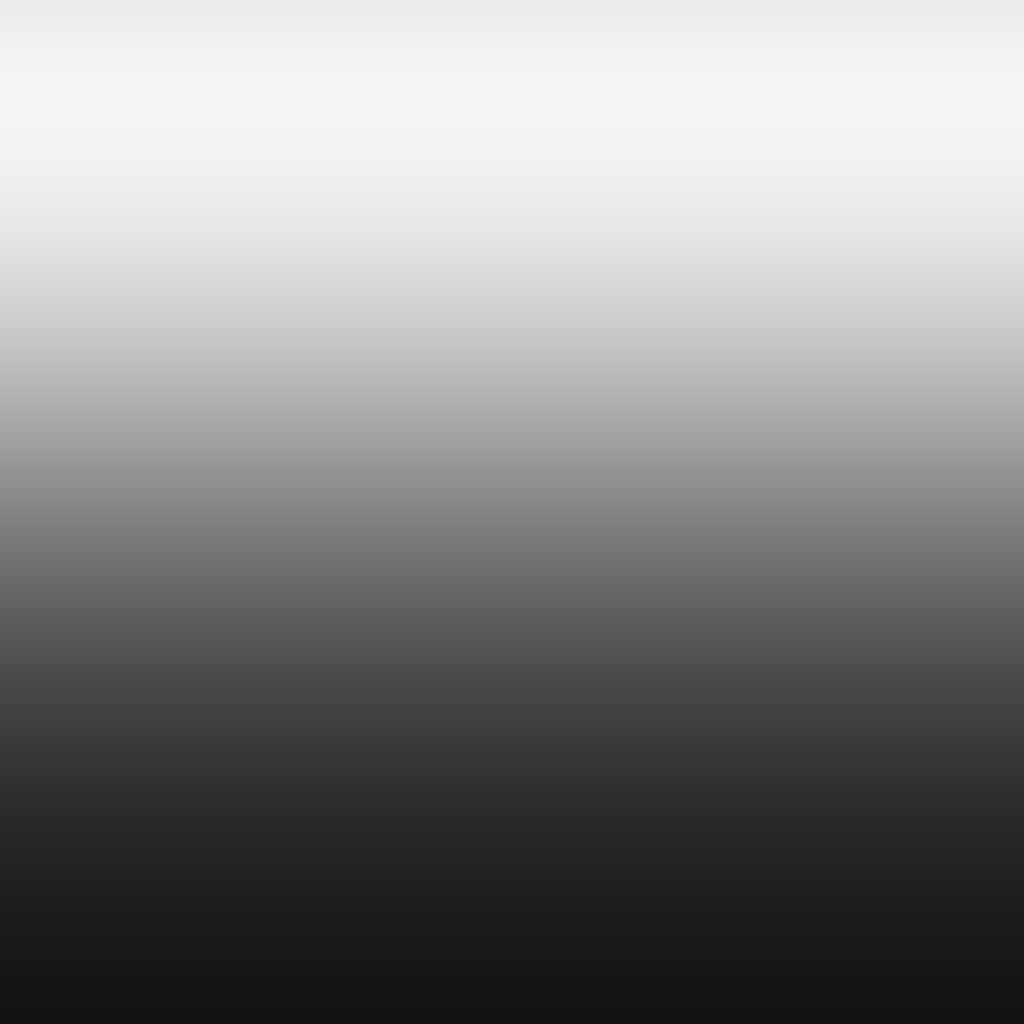} &
                \includegraphics[width=5.5mm]{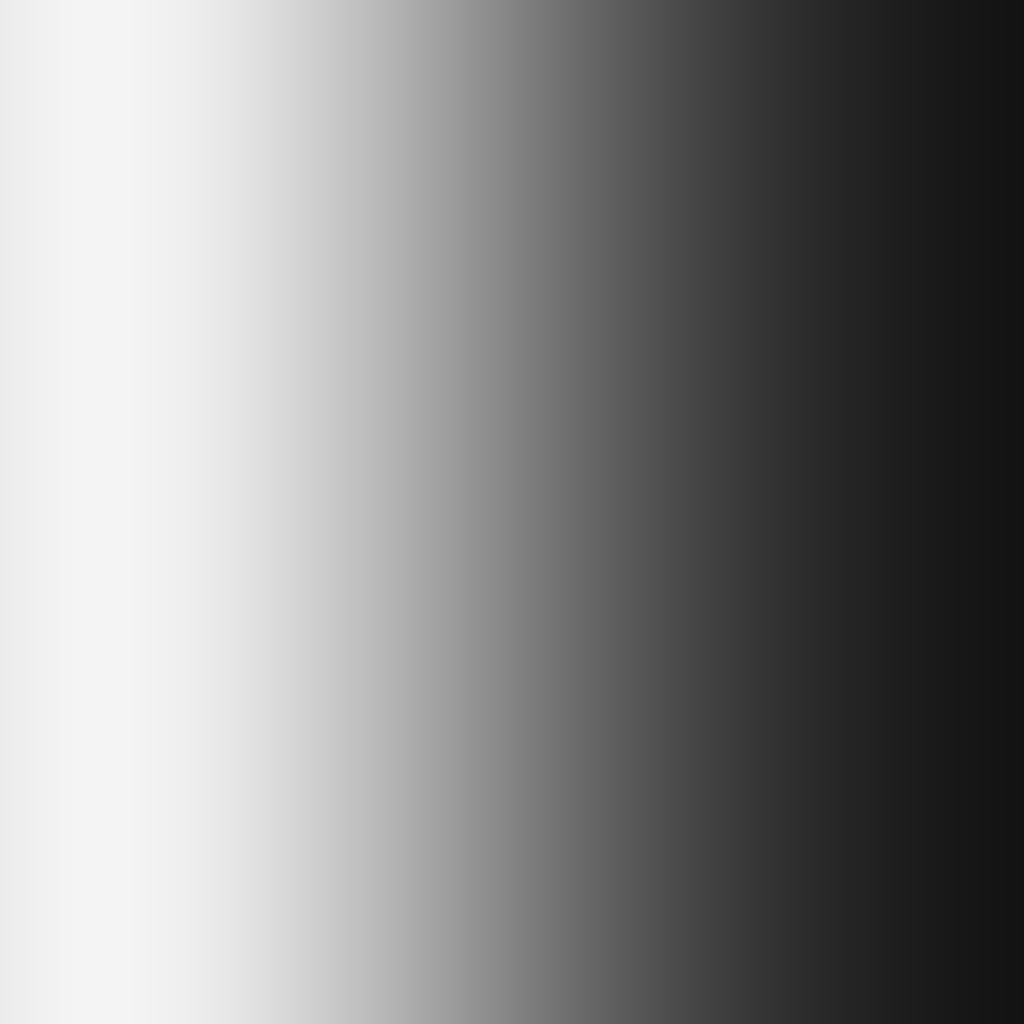} &
                \includegraphics[width=5.5mm]{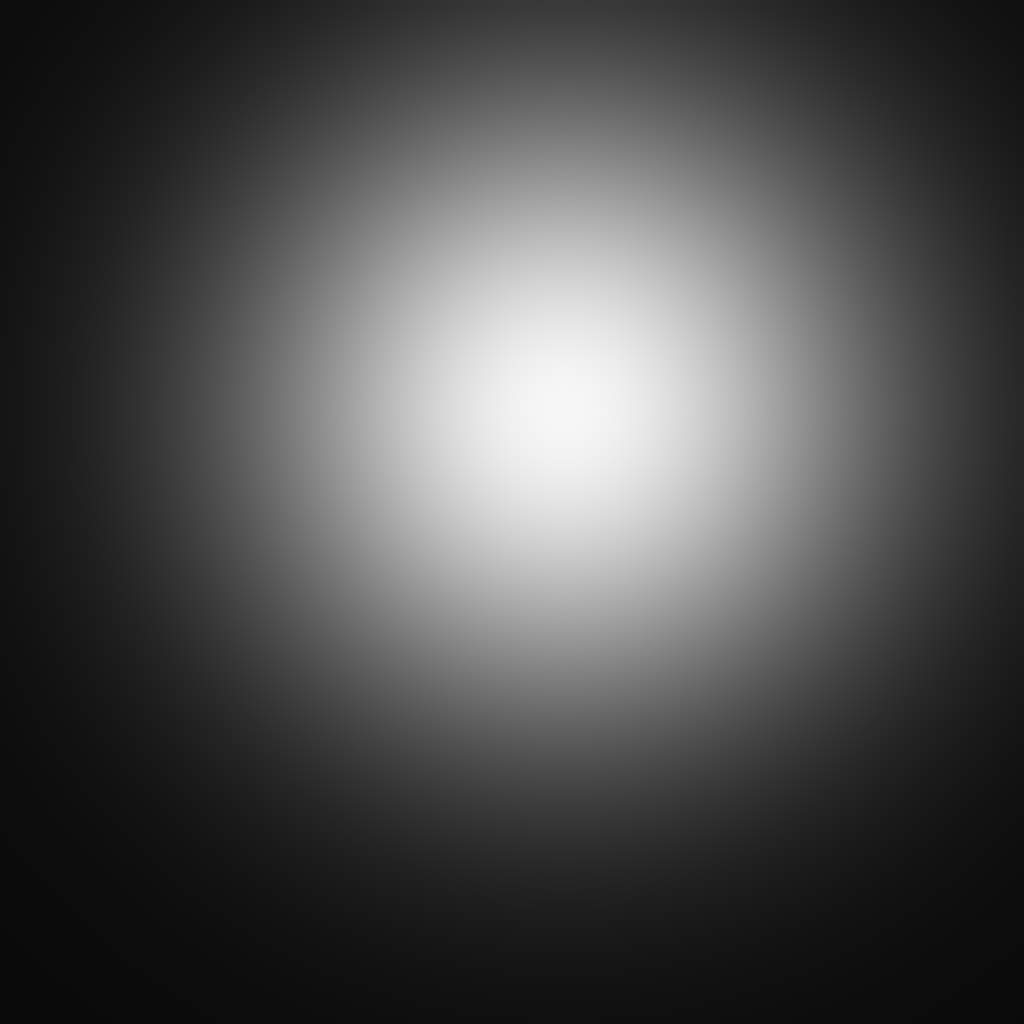} &
                \includegraphics[width=5.5mm]{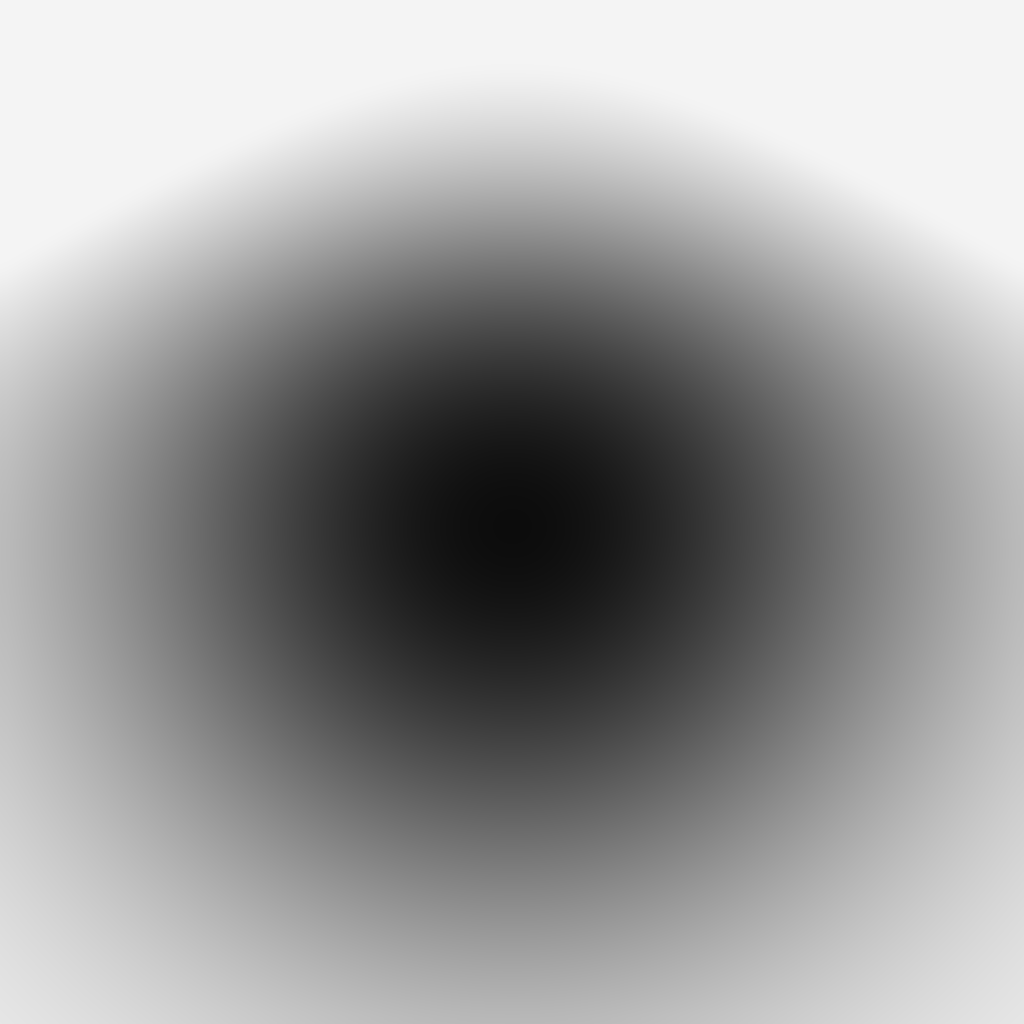} \\
                {\tiny\sffamily top} & {\tiny\sffamily side} & {\tiny\sffamily spot} & {\tiny\sffamily back} \\
                \multicolumn{4}{c}{\tiny\itshape\sffamily $+$ intensity, color temp.}
            \end{tabular}
        };
        \node[sub, below=1mm of s5] (sub5) {%
            \textendash{} illumination uniformity\\
            \textendash{} aesthetic quality\\
            \textendash{} majority vote
        };
    \end{tikzpicture}

    \vspace{1.5mm}

    \scriptsize
    \setlength{\tabcolsep}{0pt}
    \begin{tabular}{c@{\hspace{2mm}}c@{\hspace{2mm}}c@{\hspace{2mm}}c@{\hspace{2mm}}c}
        \begin{tabular}{c}
            \includegraphics[width=0.068\textwidth]{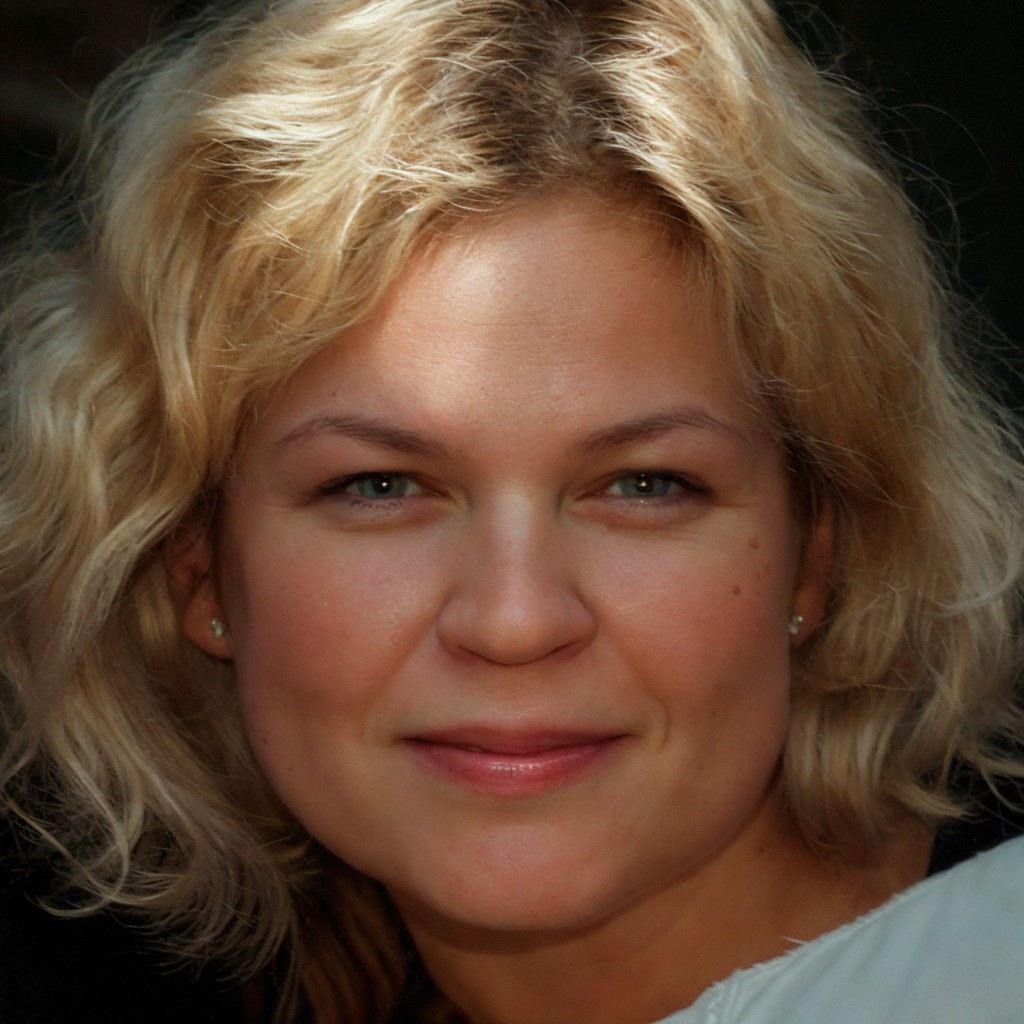} \\[0pt]
            \includegraphics[width=0.068\textwidth]{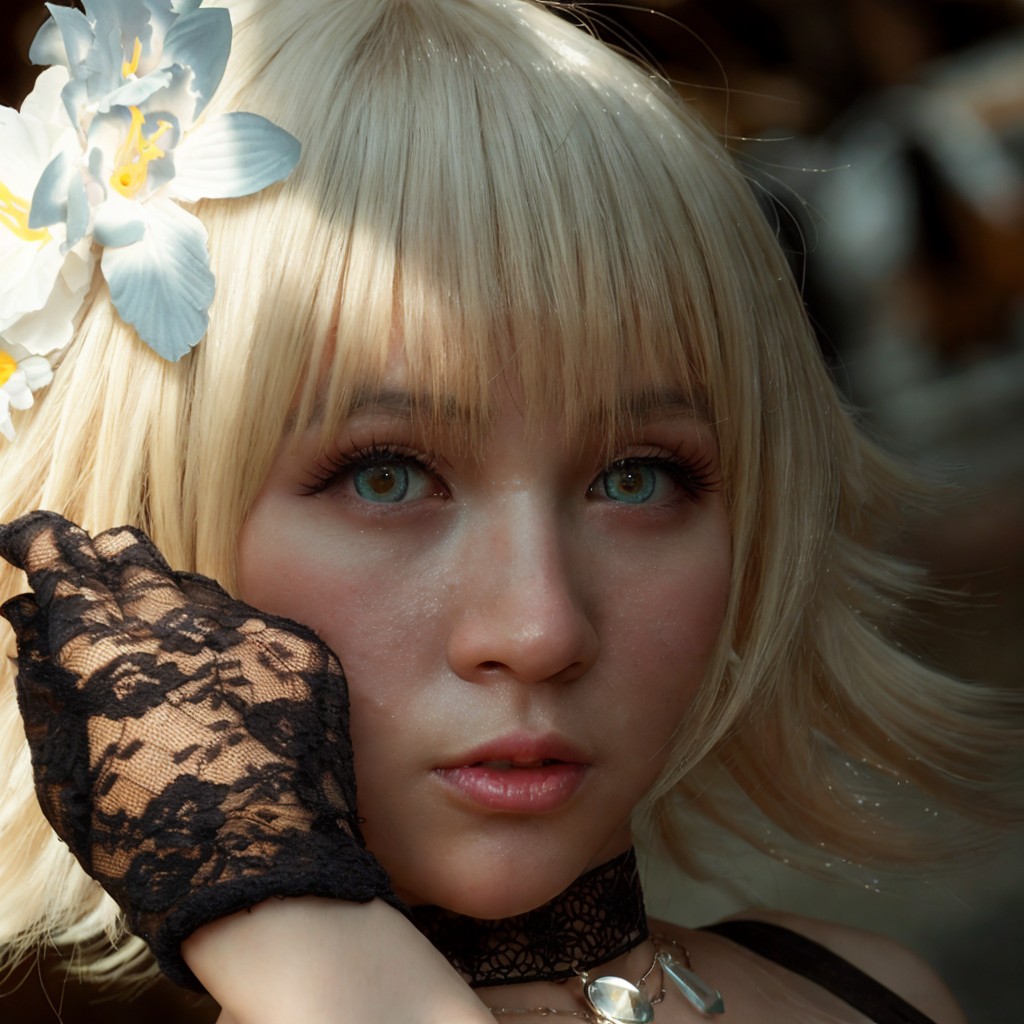}
        \end{tabular} &
        $\xrightarrow{\textit{IC-Light}}$ &
        \begin{tabular}{ccc}
            \includegraphics[width=0.065\textwidth]{Figures/DatasetSample/53919_3.jpg} &
            \includegraphics[width=0.065\textwidth]{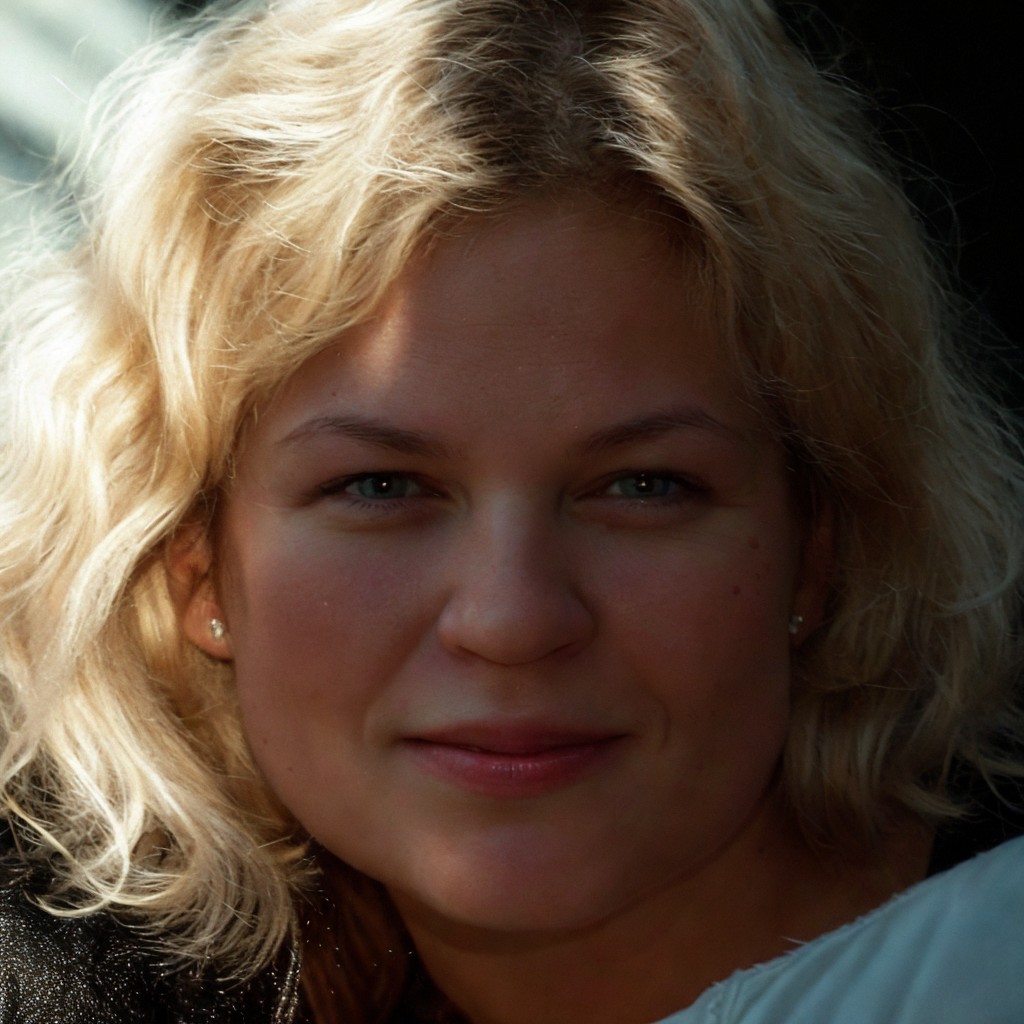} &
            \fbox{\includegraphics[width=0.065\textwidth]{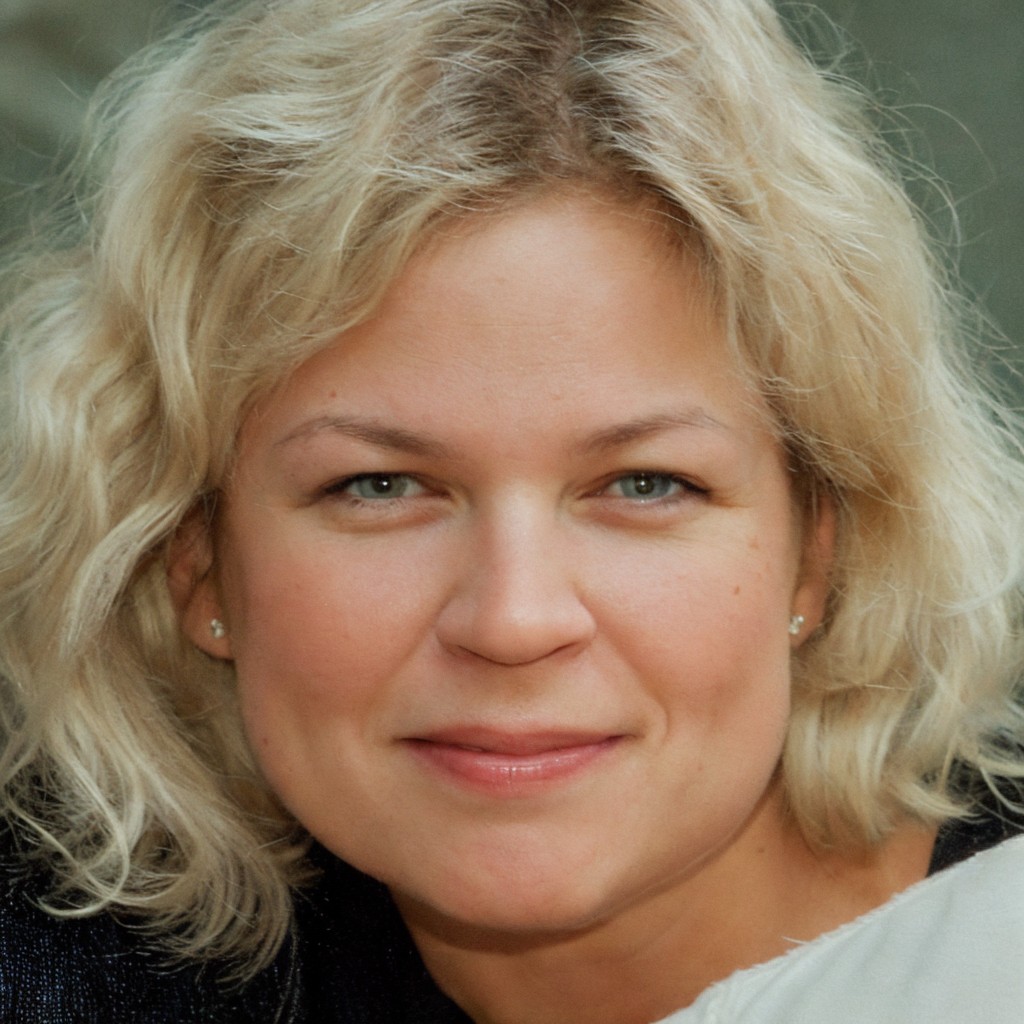}} \\[0pt]
            \includegraphics[width=0.065\textwidth]{Figures/DatasetSample/42909_0.jpg} &
            \includegraphics[width=0.065\textwidth]{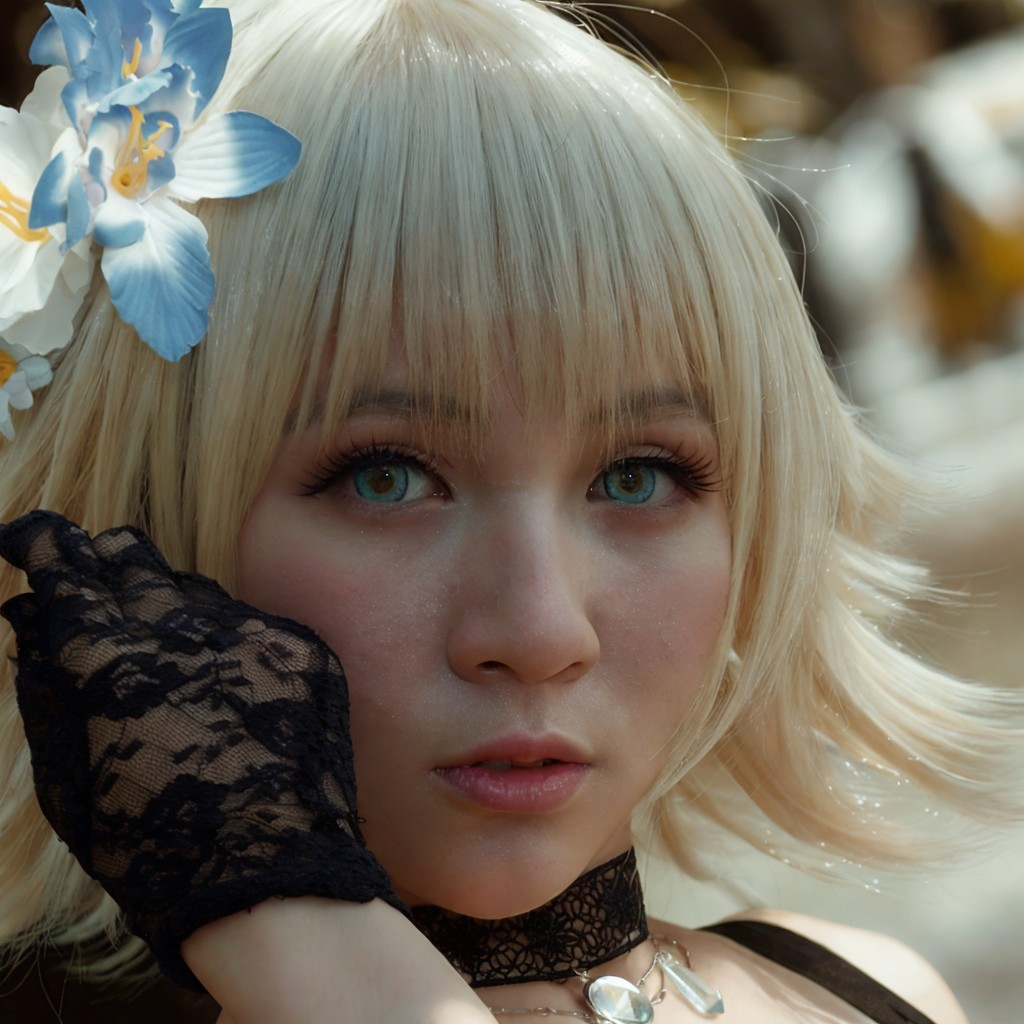} &
            \fbox{\includegraphics[width=0.065\textwidth]{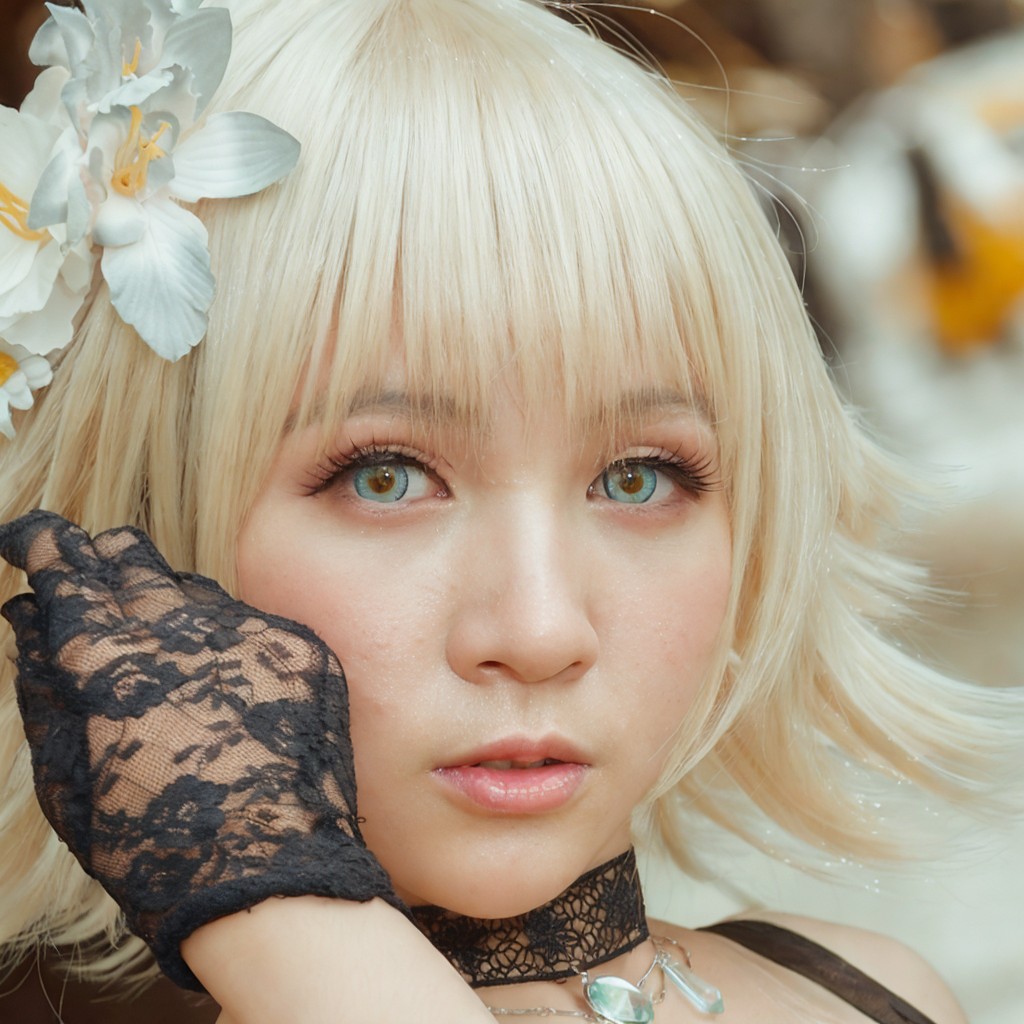}}
        \end{tabular} &
        $\xrightarrow{\textit{voting}}$ &
        \begin{tabular}{cc}
            \includegraphics[width=0.065\textwidth]{Figures/DatasetSample/53919_3.jpg} &
            \fbox{\includegraphics[width=0.065\textwidth]{Figures/DatasetSample/53919_2_GT.jpg}} \\[0pt]
            \includegraphics[width=0.065\textwidth]{Figures/DatasetSample/53919_1.jpg} &
            \fbox{\includegraphics[width=0.065\textwidth]{Figures/DatasetSample/53919_2_GT.jpg}}
        \end{tabular} \\[2pt]
        {\tiny source portraits} & & {\tiny $N$ variants per portrait (boxed: voted GT)} & & {\tiny training pairs (input, GT)} \\
    \end{tabular}

    \caption{FFHQL data generation pipeline. (Top) Abstract flow: an FFHQ portrait is fed to an extended IC-Light~\cite{zhang2025scaling} diffusion conditioned on sampled lighting controls (direction, pattern, intensity, color temperature) to produce $N$ identity-consistent variants; an expert vote then selects the most uniformly lit and aesthetically pleasing one as the GT. (Bottom) Visualization on two FFHQ identities: each source portrait yields $N$ variants (boxed image is the voted GT); each non-GT variant is paired with the same voted GT to form multiple training pairs (two such pairs shown for the first identity).}
    \label{fig:dataset}
\end{figure*}

Existing low-light datasets such as LOL~\cite{LOLDataset}, NPE~\cite{NPE}, and LIME~\cite{LIME} are limited to global lighting changes and lack fine-grained control over local illumination or identity preservation. We introduce the \textbf{Flickr-Faces-HQ-Lighting (FFHQL)} dataset, built upon FFHQ~\cite{FFHQ}, to target portrait enhancement under complex localized lighting without requiring pixel-aligned pairs.

Rather than capturing scenes under varying exposure, we leverage a generative pipeline based on IC-Light~\cite{zhang2025scaling} to synthesize diverse local lighting in direction, spatial pattern, intensity, and color temperature per portrait (Fig.~\ref{fig:dataset}). Each expert votes for one generated variant based on illumination uniformity and aesthetic quality; the majority winner is retained as the ground truth, yielding $6{,}200{+}$ identity-consistent training pairs.

\subsection{Evaluation Metrics}
\label{subsec:metric}

Because FFHQL ground truths are generated rather than pixel-aligned with the inputs, full-reference metrics (PSNR, SSIM, LPIPS) are unreliable. We therefore adopt NIQE~\cite{mittal2012niqe}, a no-reference metric based on natural-scene statistics where lower is better, as the primary quality metric and FID~\cite{heusel2017gans} for distribution-level realism. For video evaluation, we measure temporal consistency via a flow-warping error: given consecutive outputs \(O_{t-1}, O_t\) and forward optical flow \(F_{t-1\to t}\) estimated by RAFT~\cite{teed2020raft}, the per-pair error is \(\| O_t - \mathcal{W}(O_{t-1}; F_{t-1\to t}) \|_1\) over forward--backward consistent pixels. The \emph{flicker ratio} is the output warp error divided by the input warp error, with values near 1 indicating no added flicker beyond natural pixel motion.

\section{Experiments}
\label{sec:experiments}

\subsection{Experimental Setup}
\label{subsec:setup}

\textbf{Datasets.}
We train RRNet using a combined dataset of FFHQL and LOL. Without dataset-specific fine-tuning, we evaluate on six held-out low-light benchmarks: NPE~\cite{NPE}, LIME~\cite{LIME}, MEF~\cite{MEF}, DICM~\cite{DICM}, VV\footnote{https://sites.google.com/site/vonikakis/datasets}, and VCD~\cite{VCD}, which consists of video-conferencing scenarios.

\textbf{Implementation.}
We train with the four losses in Sec.~\ref{subsec:lossfunc} using Adam (lr $1{\times}10^{-4}$) for 300K iterations on a single RTX 3090. Depth maps $M$ come from a frozen DepthAnything-Small~\cite{depth_anything_v2}.

\subsection{Comparison with State-of-the-Art Methods}
\label{subsec:comparison}

We compare RRNet with representative state-of-the-art image and video enhancement methods.
For image enhancement, we include EnlightenGAN~\cite{Jiang19}, RetinexNet~\cite{Wei18}, DeepUPE~\cite{wang2019deepupe}, Zero-DCE++~\cite{Zero-DCE++}, SNRNet~\cite{snrnet}, PairLIE~\cite{PairLIE}, RetinexMamba~\cite{retinexmambabai2024}, CWNet~\cite{cwnetzhang2025cwnet_iccv}, HVI-CIDNet~\cite{cidnetyan2025hvi}, DarkIR~\cite{feijoo2025darkir}, GLARE~\cite{zhou2024glare}, and ReDDiT~\cite{lan2025reddit}.
For video enhancement, we compare with FastLLVE~\cite{fastllve} and StableLLVE~\cite{StableLLVE}.

\textbf{VisualComparison.}
Fig.~\ref{fig:visual_comparison} shows three challenging unbalanced-exposure cases from FFHQ, VV, and VCD; RRNet performs spatially adaptive correction, preserves natural skin tones, and reduces artifacts by operating on virtual lighting parameters rather than pixel intensities.

\begin{table}[!t]
    \centering
    \caption{NIQE comparison on various datasets. Lower NIQE scores indicate higher quality. \textbf{Throughout all tables in this paper, bold marks the best result and underline marks the second-best.} Methods listed in the lower half of the table are expected to achieve real-time performance during testing. Rows marked $^\dagger$ are values reported by the original publications; ``---'' indicates a benchmark not reported.}
    \label{tab:comparison}
    \footnotesize
    \setlength{\tabcolsep}{3pt}
    \renewcommand{\arraystretch}{1.2}
    \begin{tabular}{lcccccccc}
        \hline
        \textbf{Method} & \textbf{NPE} & \textbf{LIME} & \textbf{MEF} & \textbf{VV} & \textbf{DICM} & \textbf{VCD} & \textbf{Avg.} & \textbf{Portrait} \\
        \hline
        EnlightenGAN & 4.11 & \textbf{3.72} & \underline{3.32} & 2.68 & \textbf{3.45} & 4.74 & \underline{3.67} & \underline{2.63} \\
        DeepUPE & 3.67 & 4.14 & 3.68 & 3.22 & 3.89 & 5.13 & 3.96 & 3.14 \\
        RetinexNet & 4.46 & 4.42 & 3.98 & 2.98 & 4.20 & \textbf{3.68} & 3.95 & 2.95 \\
        PairLIE & 4.02 & 4.51 & 4.16 & 3.66 & 4.09 & 4.56 & 4.08 & 3.53 \\
        RetinexMamba & 3.55 & 3.88 & \underline{3.32} & 5.62 & 3.57 & 5.28 & 4.20 & 5.57 \\
        CWNet & 3.65 & 4.46 & 4.38 & \underline{2.62} & 3.83 & 5.12 & 4.01 & 2.64 \\
        CIDNet & 3.74 & 3.81 & 3.34 & 3.21 & 3.78 & 4.84 & 3.79 & 3.15 \\
        DarkIR-m$^\dagger$ & 3.99 & 4.07 & 3.45 & 3.74 & 3.76 & --- & 3.80 & --- \\
        GLARE$^\dagger$ & 4.19 & 4.52 & 3.66 & --- & 3.61 & --- & 3.75 & --- \\
        ReDDiT$^\dagger$ & \textbf{3.24} & 3.45 & 3.93 & 3.00 & 3.62 & --- & 3.65 & --- \\
        \hline
        Zero-DCE++ & \underline{3.47} & 3.97 & 3.40 & 3.10 & \underline{3.54} & 4.85 & 3.72 & 3.00 \\
        SNRNet & 4.32 & 5.74 & 4.18 & 6.87 & 4.10 & 9.02 & 5.71 & 9.23 \\
        FastLLVE & 4.76 & 5.19 & 5.69 & 4.25 & 5.55 & 5.29 & 5.12 & 4.23 \\
        StableLLVE & 3.62 & 4.22 & 3.92 & 3.20 & 3.83 & 5.11 & 3.82 & 3.82 \\
        \hline
        RRNet (Ours) & 3.62 & \underline{3.75} & \textbf{3.24} & \textbf{2.50} & 3.83 & \underline{4.26} & \textbf{3.53} & \textbf{2.44} \\
        \hline
    \end{tabular}
\end{table}

\textbf{Quantitative Comparison.} Table~\ref{tab:comparison} reports NIQE across six benchmarks and a portrait subset (lower is better). RRNet attains the best average ($3.53$) and portrait ($2.44$) scores and is top-two on four benchmarks. Among methods evaluated on all six, it is the only one whose worst-case NIQE stays below $4.30$ ($4.26$ on VCD), whereas every other peaks above $4.40$ and most above $4.70$ (e.g., RetinexMamba $5.62$ on VV, SNRNet $9.02$ on VCD); the $\dagger$ methods omit VCD, the hardest set. This worst-case-bounded stability is what a deployed enhancer requires.

\subsection{Ablation Study}
\label{subsec:ablation}

Tables~\ref{tab:ablation_structure}--\ref{tab:ablation_loss} validate each design choice. The dual-branch LPRM and AGM both improve NIQE; $K=9$ virtual lights is Pareto-optimal — fewer cannot capture complex local lighting, more are redundant. The Retinex consistency loss further improves both NIQE and FID over the regularized pixel+ROI baseline.

\begin{table}[!t]
    \centering
    \caption{Ablation study on model architecture (FFHQL dataset). Baseline: dual-branch LPRM with AGM and 9 virtual lights. Time is measured on an NVIDIA RTX 3090 GPU.}
    \label{tab:ablation_structure}
    \footnotesize
    \setlength{\tabcolsep}{4pt}
    \renewcommand{\arraystretch}{1.15}
    \begin{tabular}{lccc}
        \hline
        \textbf{Modification} & \textbf{NIQE $\downarrow$} & \textbf{FID $\downarrow$} & \textbf{Time (ms) $\downarrow$} \\
        \hline
        None (Baseline) & \textbf{3.69} & \textbf{22.84} & 17.0 \\
        Single-branch & 4.00 & 23.35 & 9.5 \\
        Remove AGM & 4.05 & 23.03 & 15.7 \\
        Single-branch + Remove AGM & 4.06 & 23.51 & 8.9 \\
        \hline
    \end{tabular}
\end{table}

\begin{table}[!t]
    \centering
    \caption{Ablation study on number of virtual lights (FFHQL validation set).}
    \label{tab:ablation_lights}
    \footnotesize
    \setlength{\tabcolsep}{4pt}
    \renewcommand{\arraystretch}{1.15}
    \begin{tabular}{lccc}
        \hline
        \textbf{\# Lights} & \textbf{NIQE $\downarrow$} & \textbf{FID $\downarrow$} & \textbf{Time (ms) $\downarrow$} \\
        \hline
        3  & 3.78          & 23.71          & 16.5 \\
        6  & 3.71          & 22.97          & 16.7 \\
        9  & \textbf{3.69} & \textbf{22.84} & 17.0 \\
        12 & 3.72          & 24.25          & 17.5 \\
        \hline
    \end{tabular}
    \vspace{2pt}
\end{table}

\begin{table}[!t]
    \centering
    \caption{Ablation study on loss functions (FFHQL dataset).}
    \label{tab:ablation_loss}
    \footnotesize
    \setlength{\tabcolsep}{6pt}
    \renewcommand{\arraystretch}{1.15}
    \begin{tabular}{lcc}
        \hline
        \textbf{Loss Configuration} & \textbf{NIQE $\downarrow$} & \textbf{FID $\downarrow$} \\
        \hline
        \(\mathcal{L}_{\text{pixel}}\) only & 3.78 & 23.01 \\
        \(\mathcal{L}_{\text{pixel}} + \mathcal{L}_{\text{roi}}\) & 3.72 & 24.69 \\
        \(\mathcal{L}_{\text{pixel}} + \mathcal{L}_{\text{roi}} + \lambda_r \mathcal{L}_{\text{reg}}\) & 3.71 & 23.05 \\
        Full (\(+ \lambda_{\text{ret}} \mathcal{L}_{\text{retinex}}\)) & \textbf{3.69} & \textbf{22.84} \\
        \hline
    \end{tabular}
    \vspace{2pt}
\end{table}

\subsection{Video Processing}
\label{subsec:video}

\begin{figure}[htbp]
    \centering
    \setlength{\tabcolsep}{1pt}
    \begin{tabular}{ccc}
        \includegraphics[width=0.30\columnwidth]{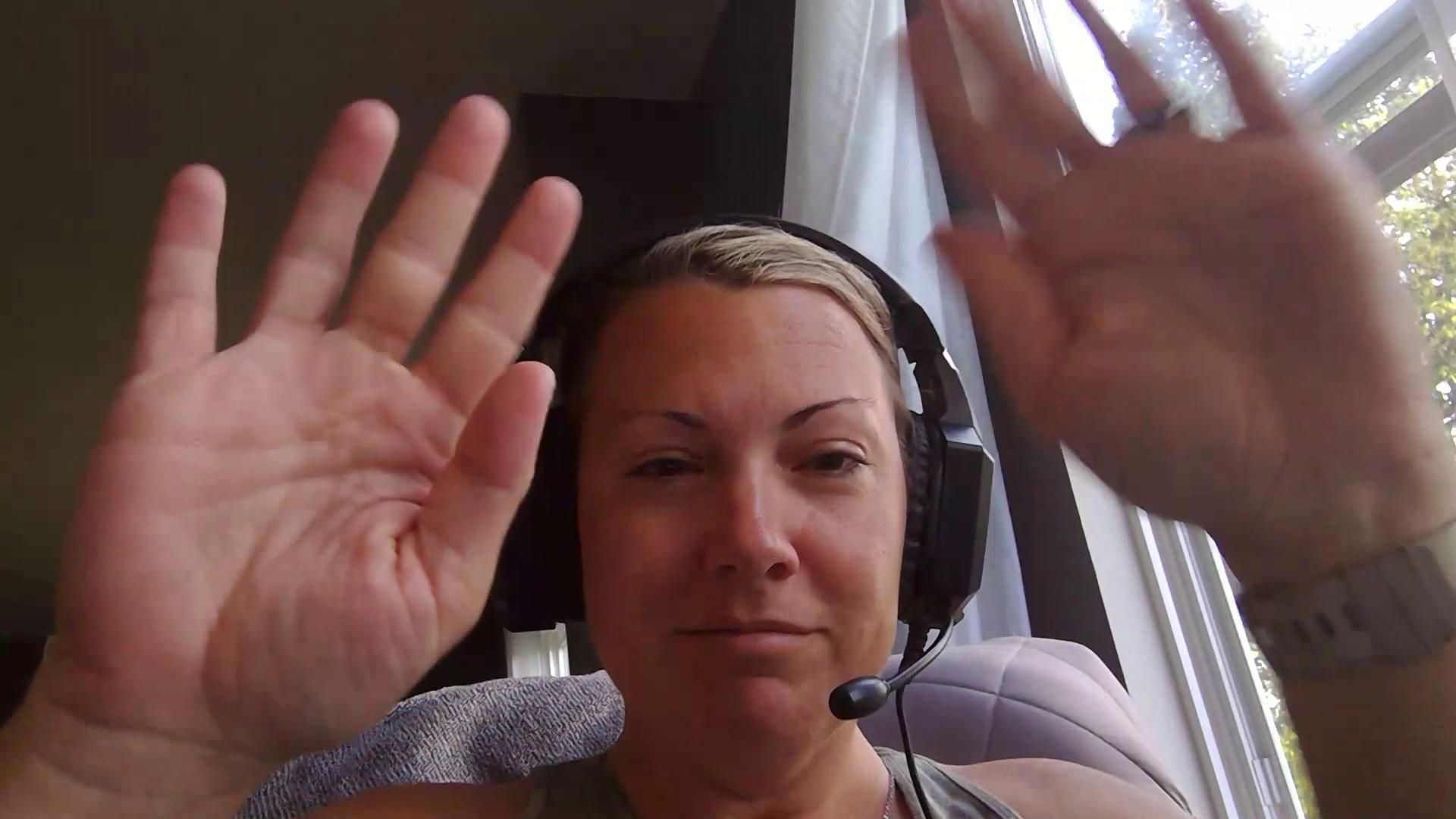} &
        \includegraphics[width=0.30\columnwidth]{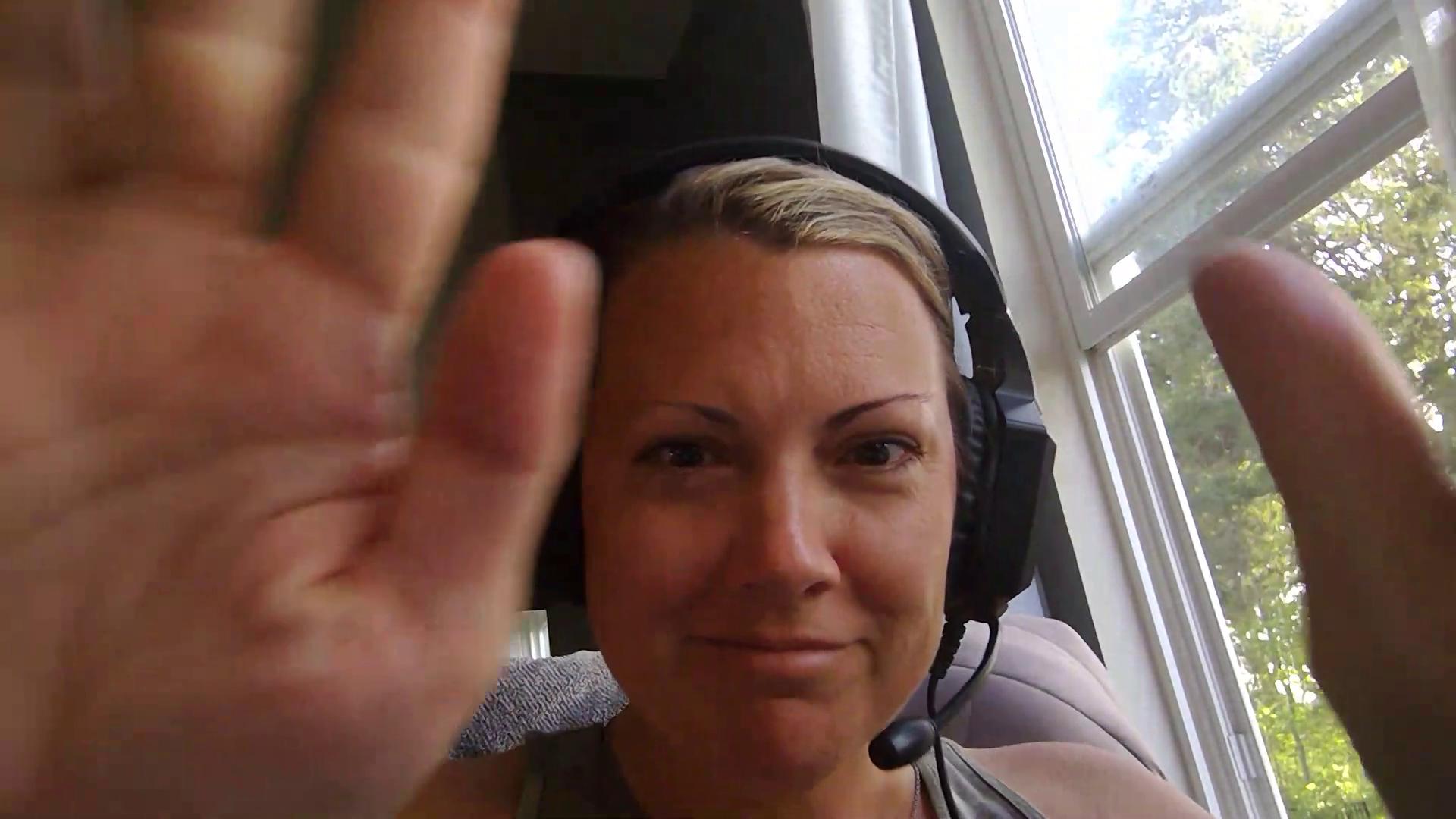} &
        \includegraphics[width=0.30\columnwidth]{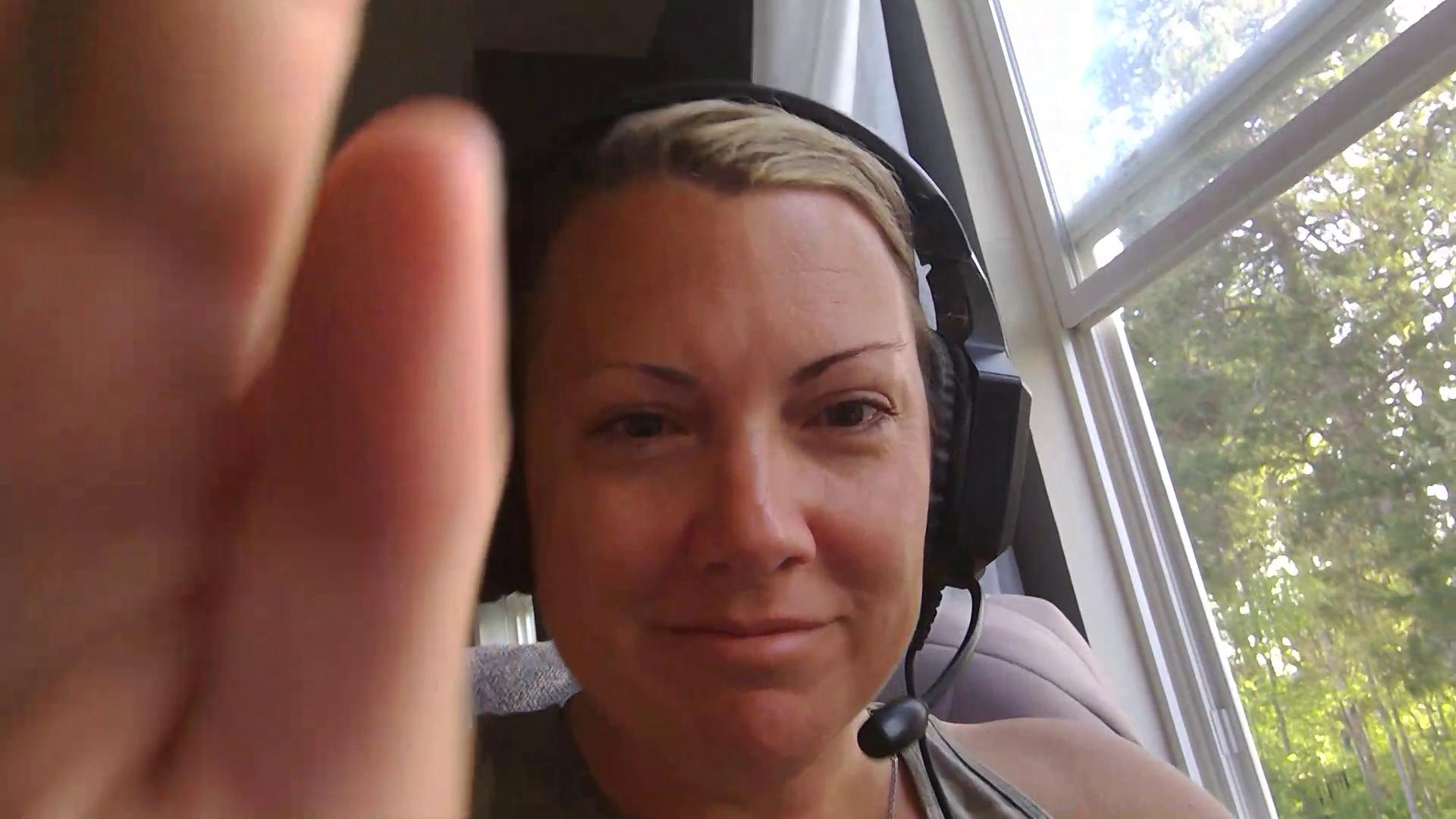} \\
        \multicolumn{3}{c}{\includegraphics[width=0.80\columnwidth]{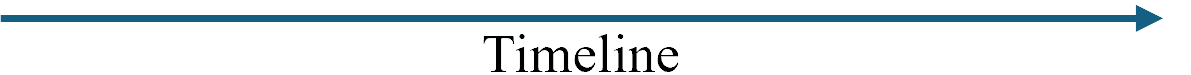}} \\
        \includegraphics[width=0.30\columnwidth]{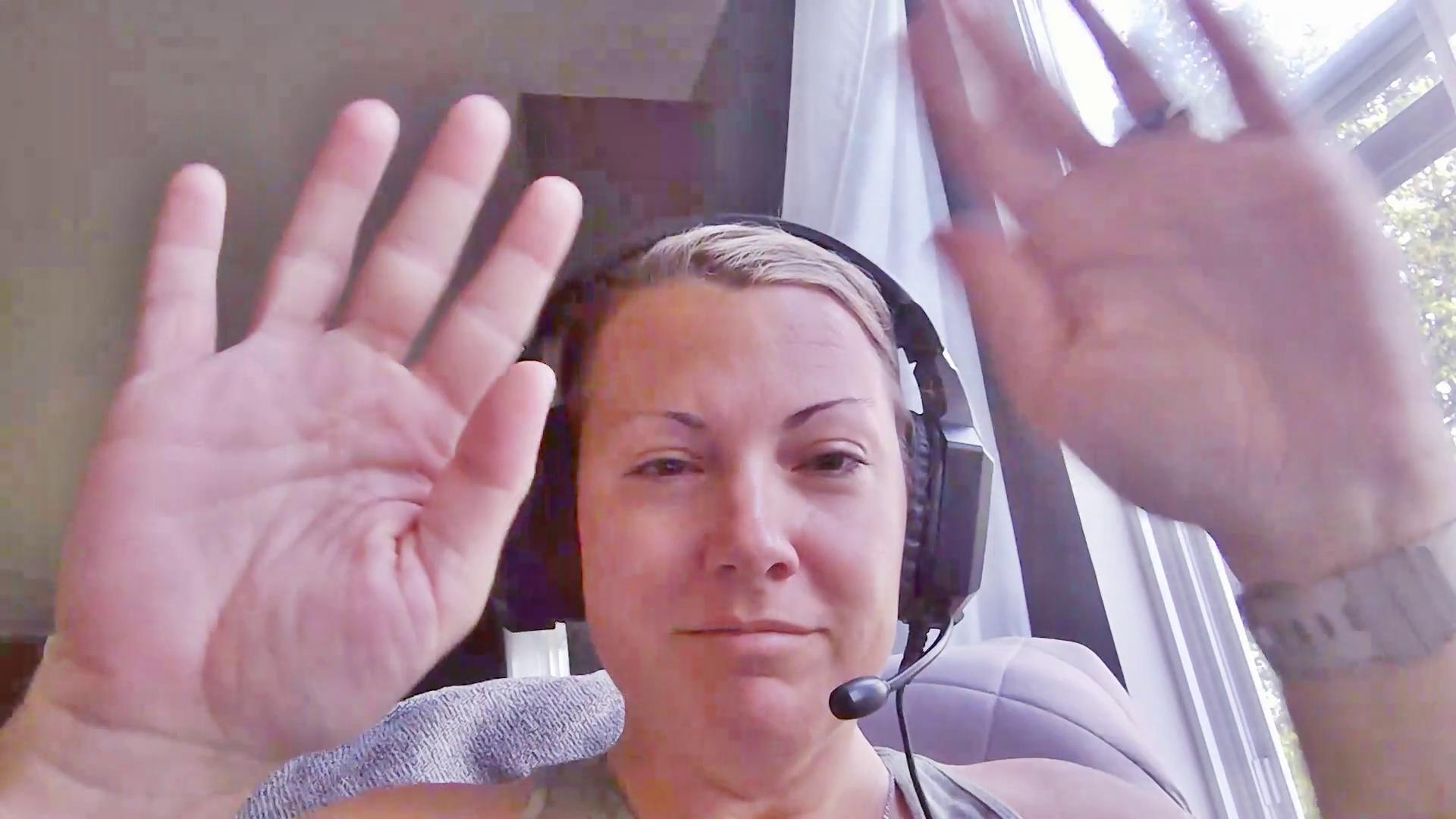} &
        \includegraphics[width=0.30\columnwidth]{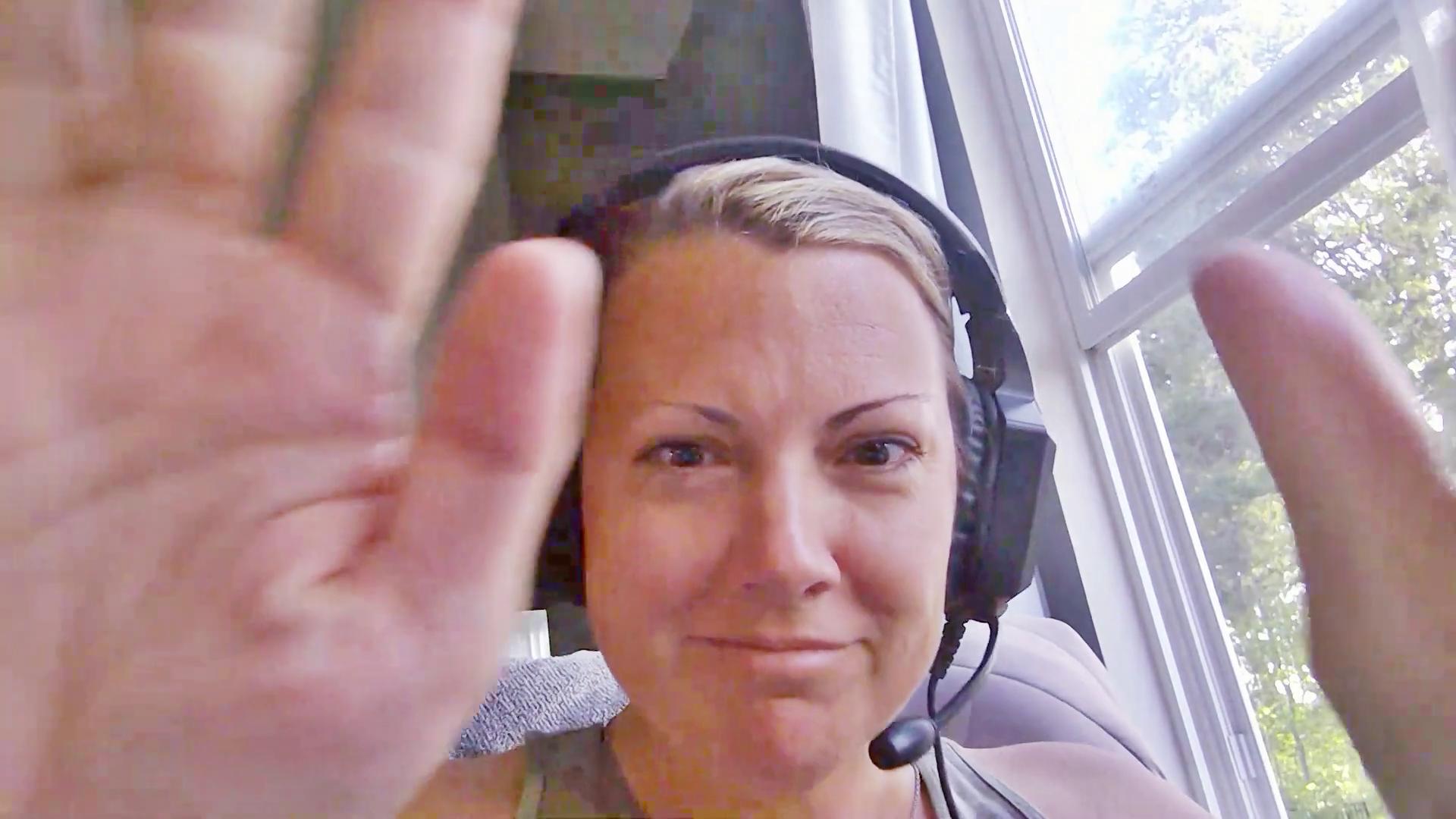} &
        \includegraphics[width=0.30\columnwidth]{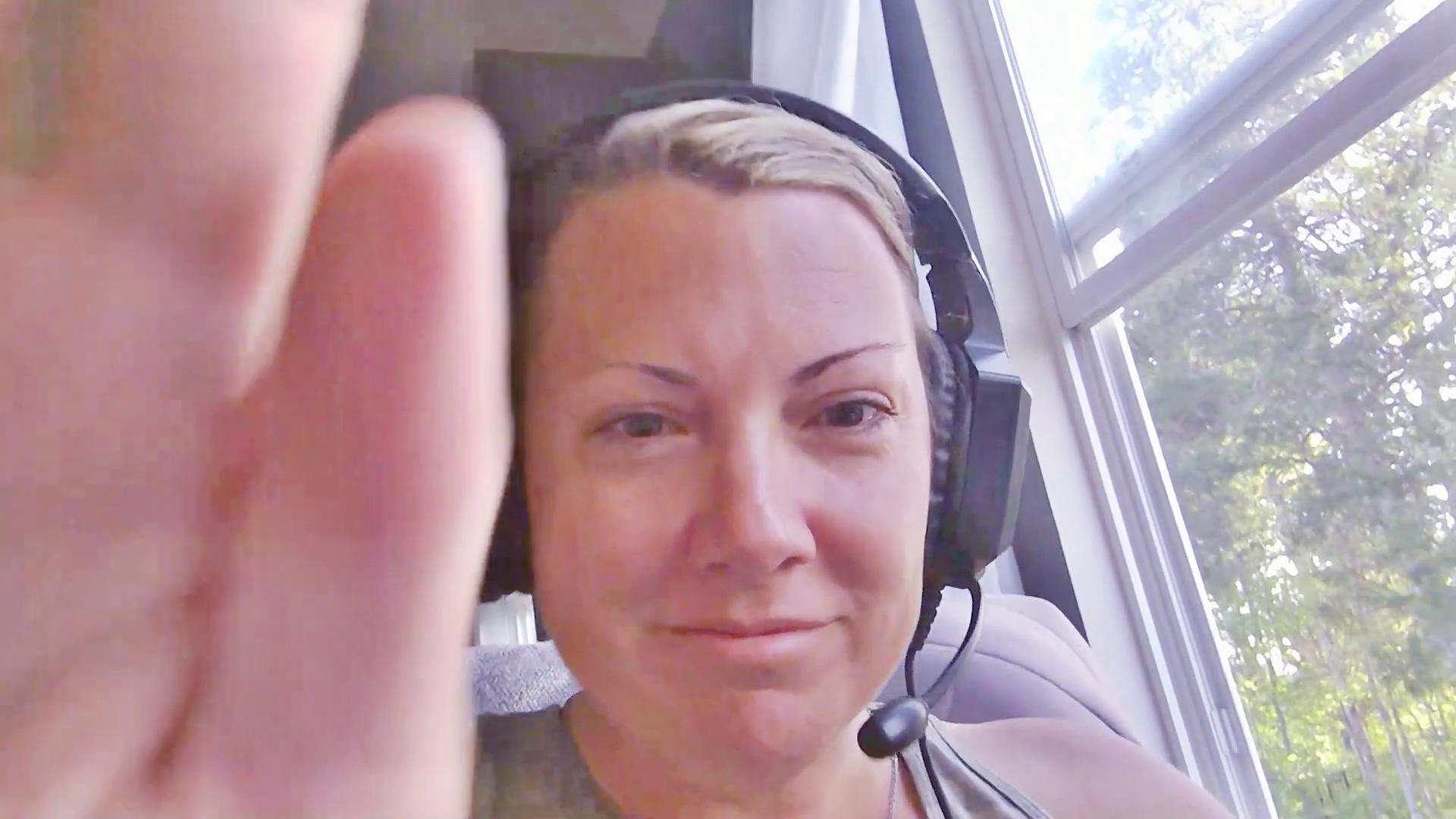} \\
        \includegraphics[width=0.30\columnwidth]{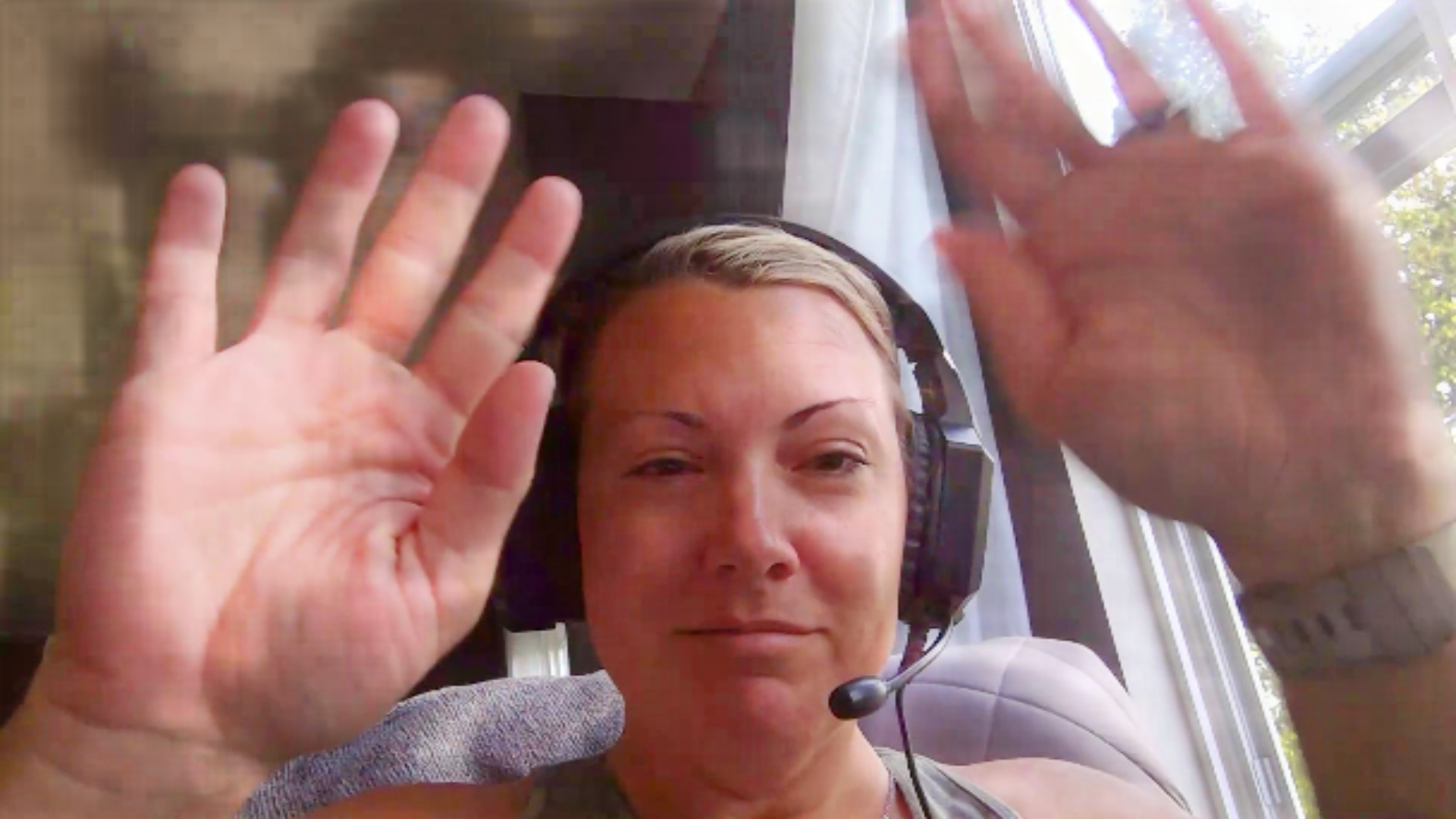} &
        \includegraphics[width=0.30\columnwidth]{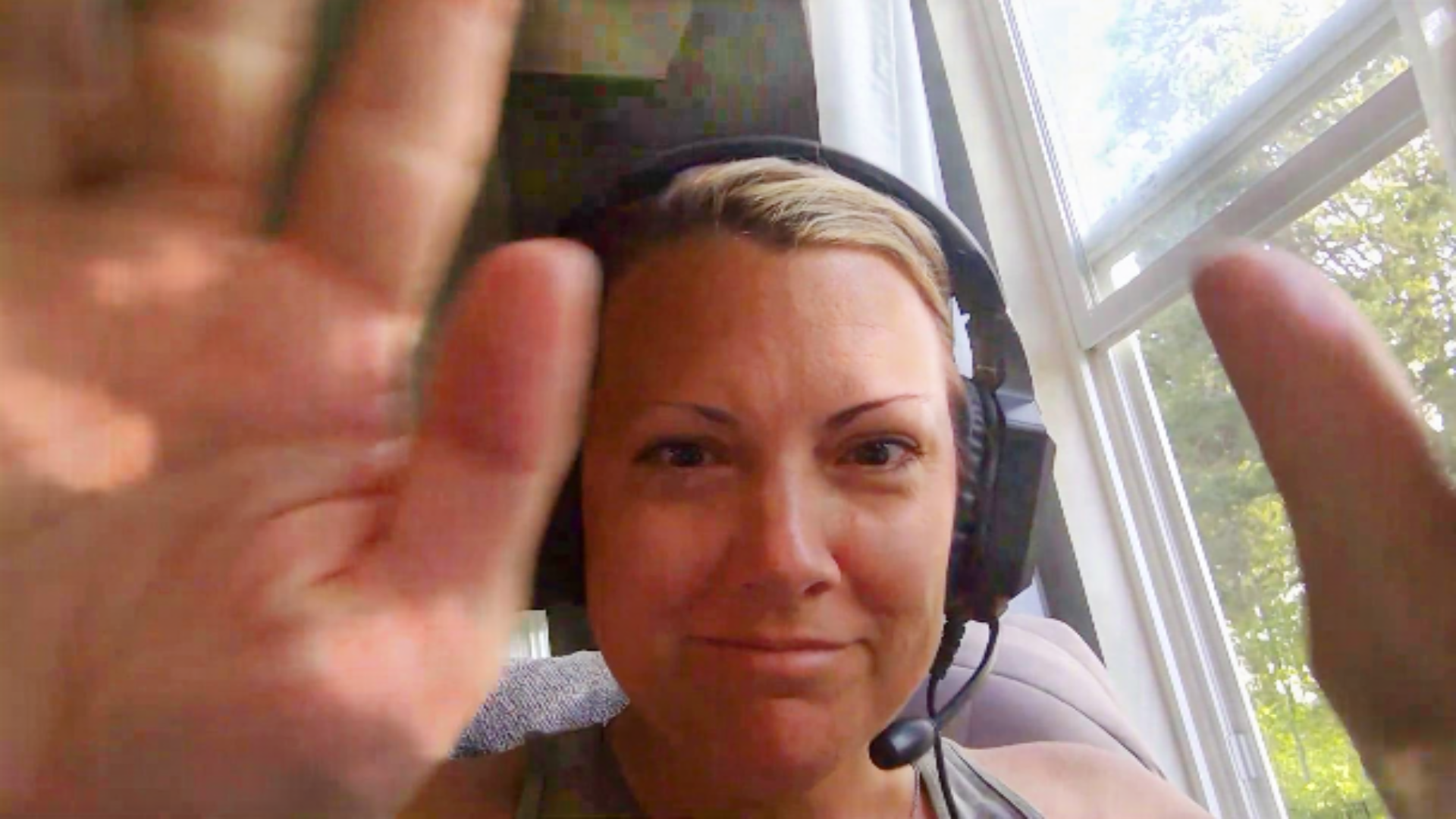} &
        \includegraphics[width=0.30\columnwidth]{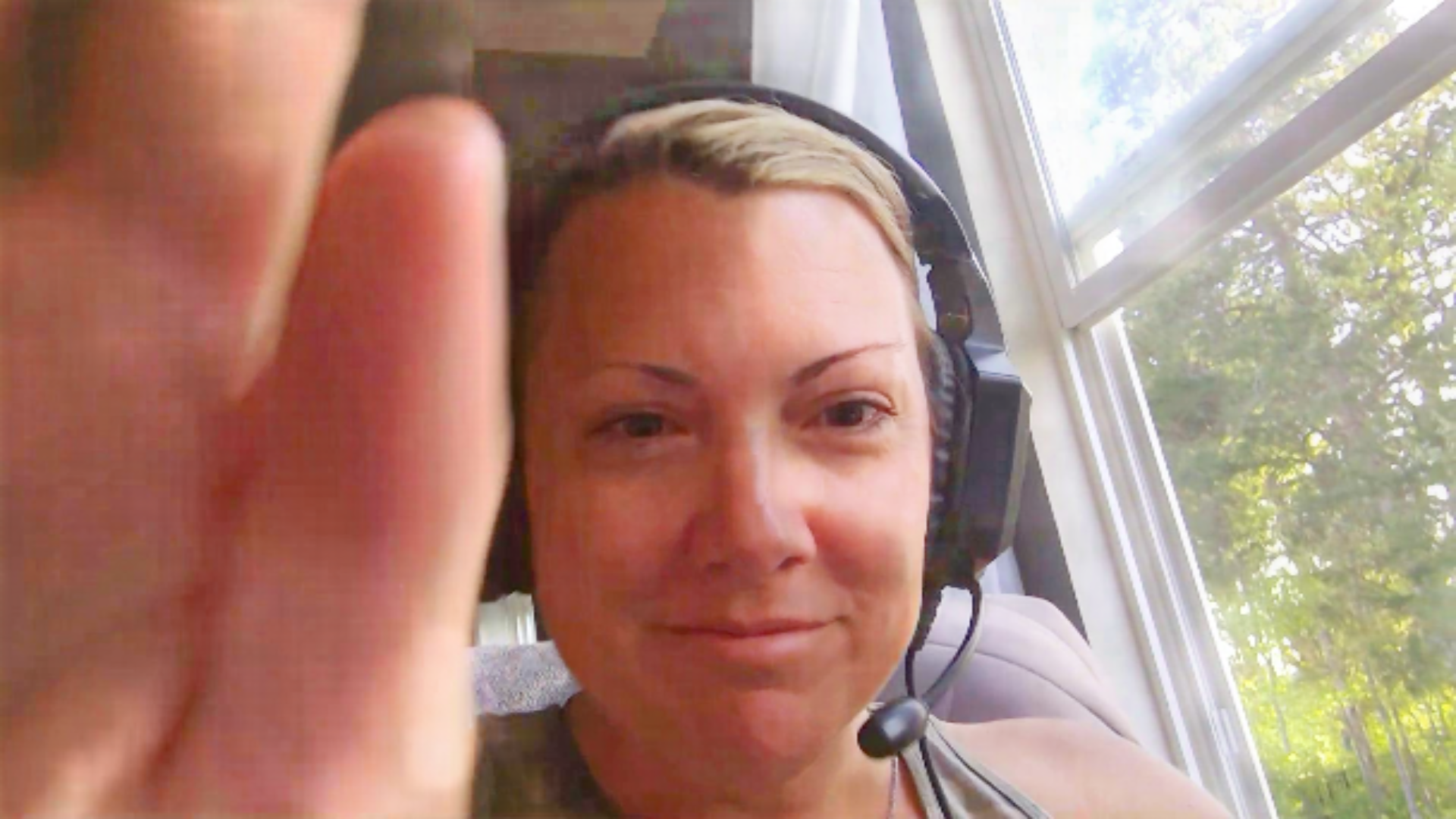} \\
        \includegraphics[width=0.30\columnwidth]{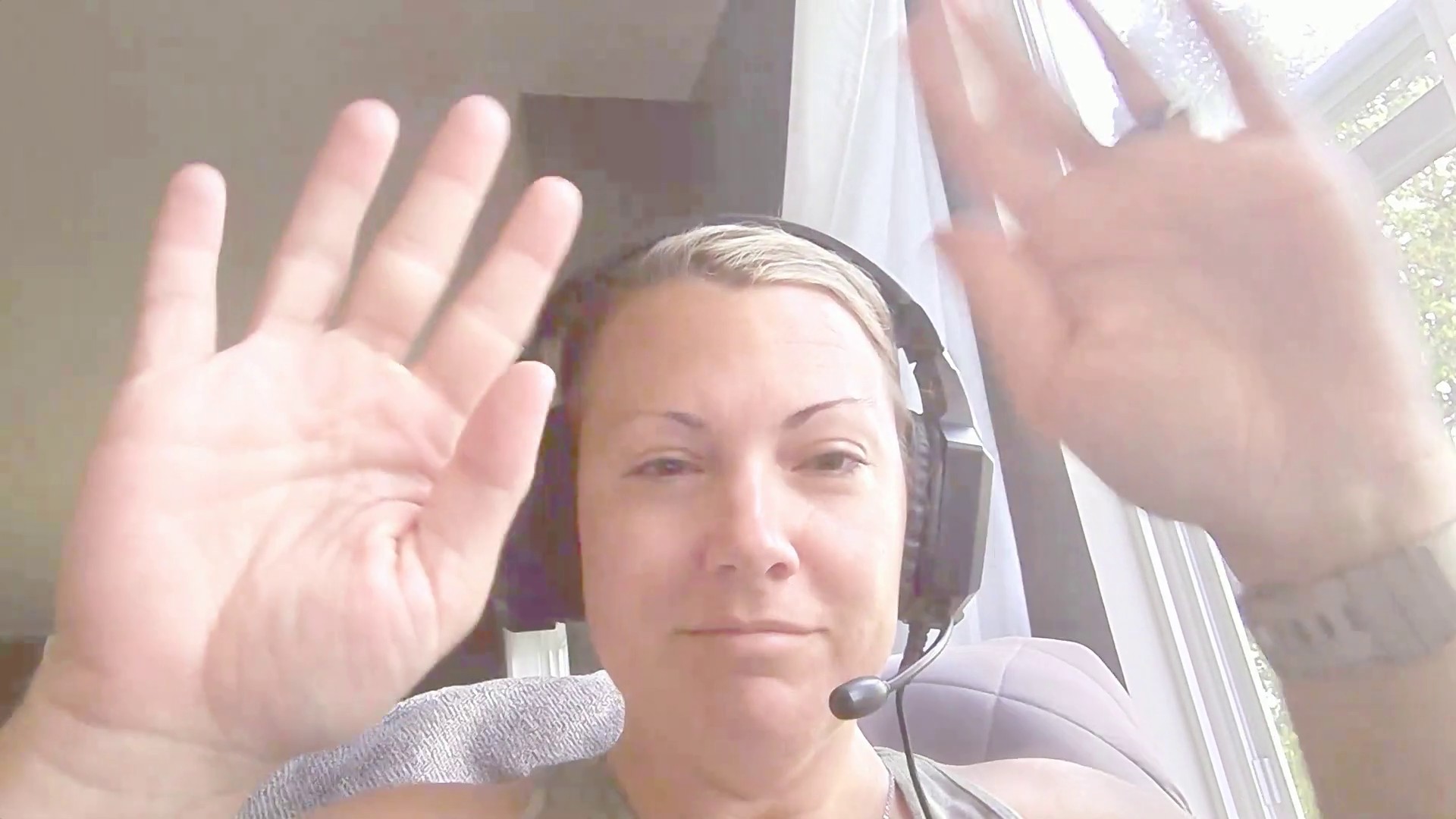} &
        \includegraphics[width=0.30\columnwidth]{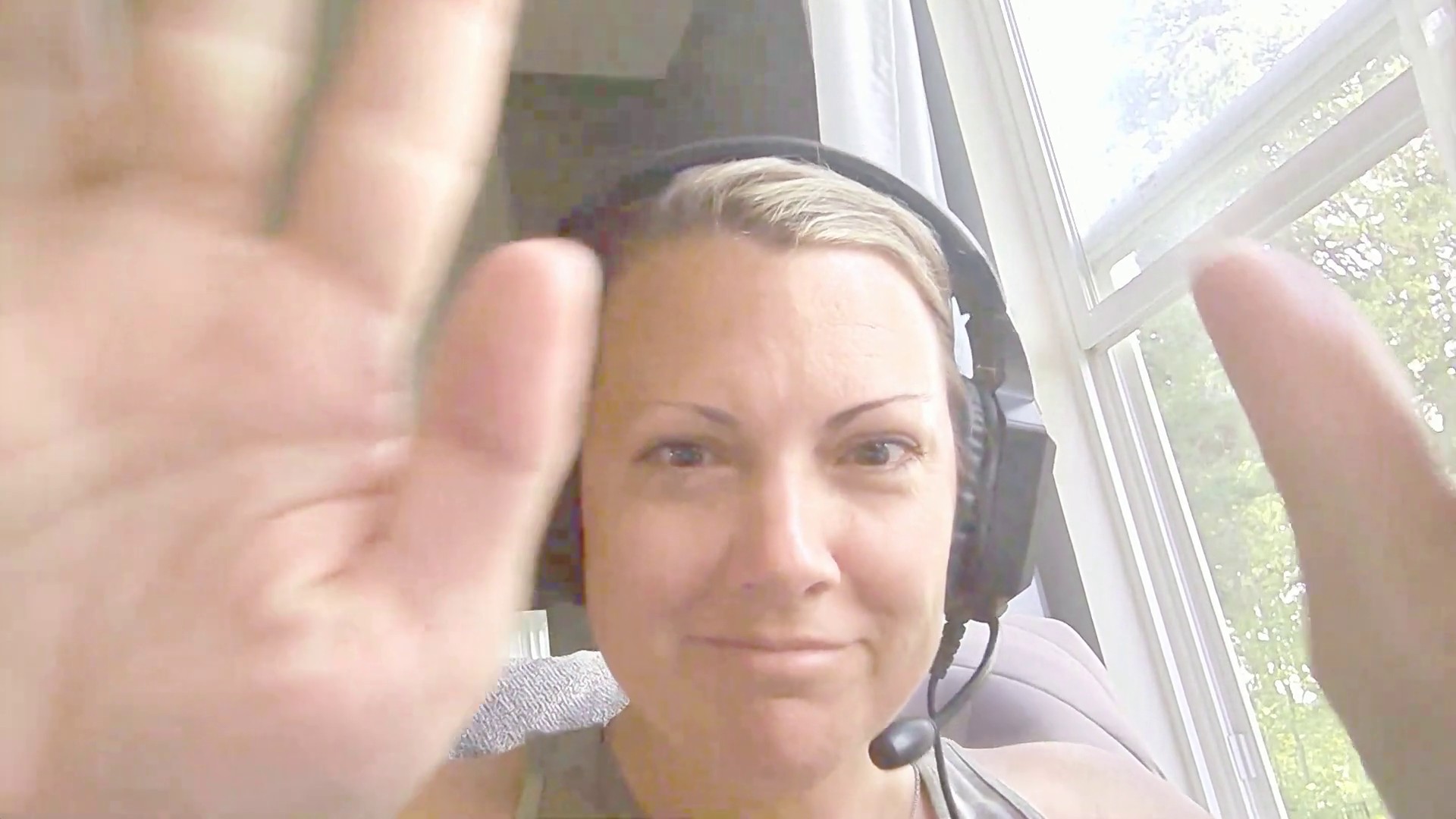} &
        \includegraphics[width=0.30\columnwidth]{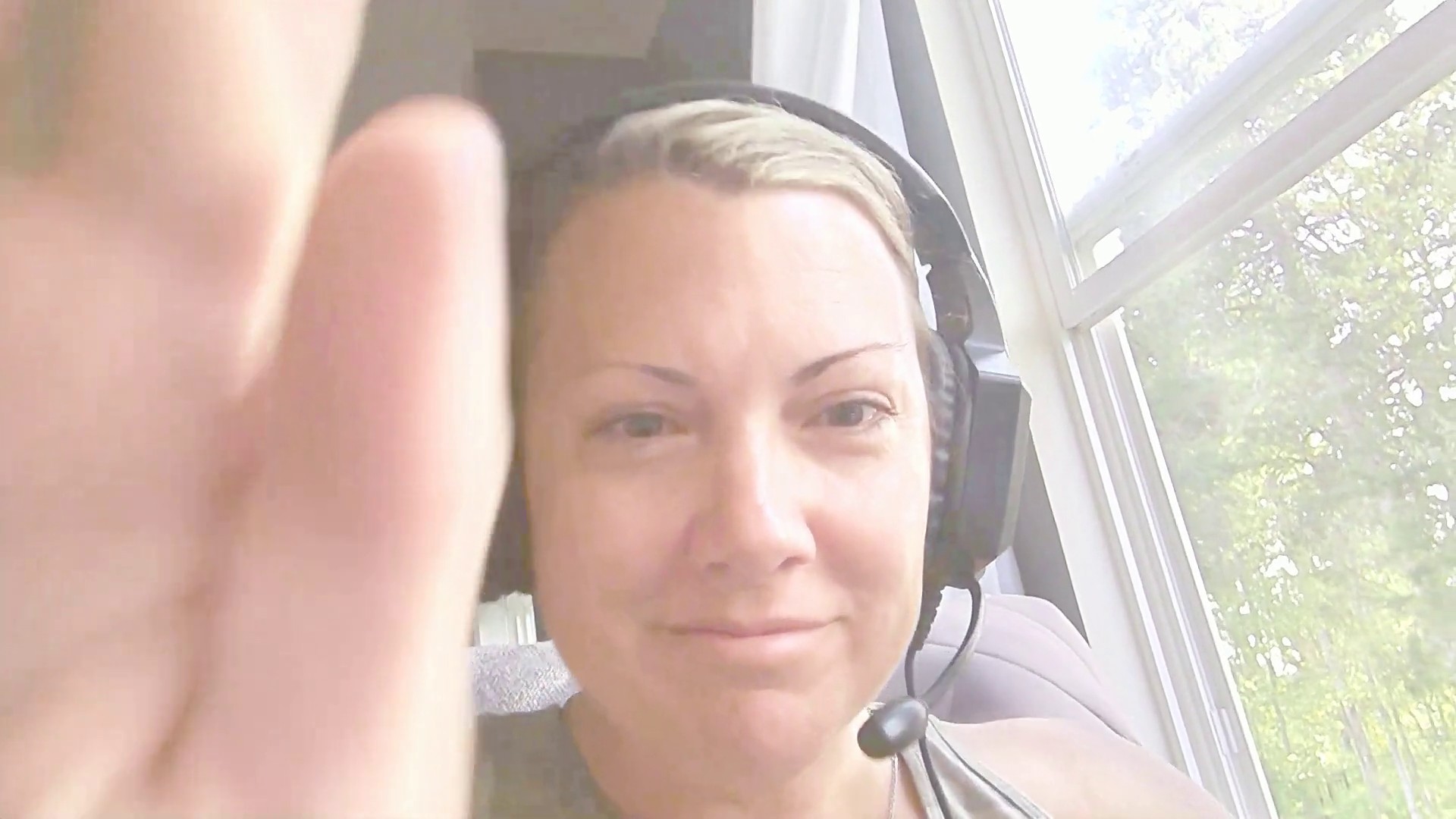} \\
        \includegraphics[width=0.30\columnwidth]{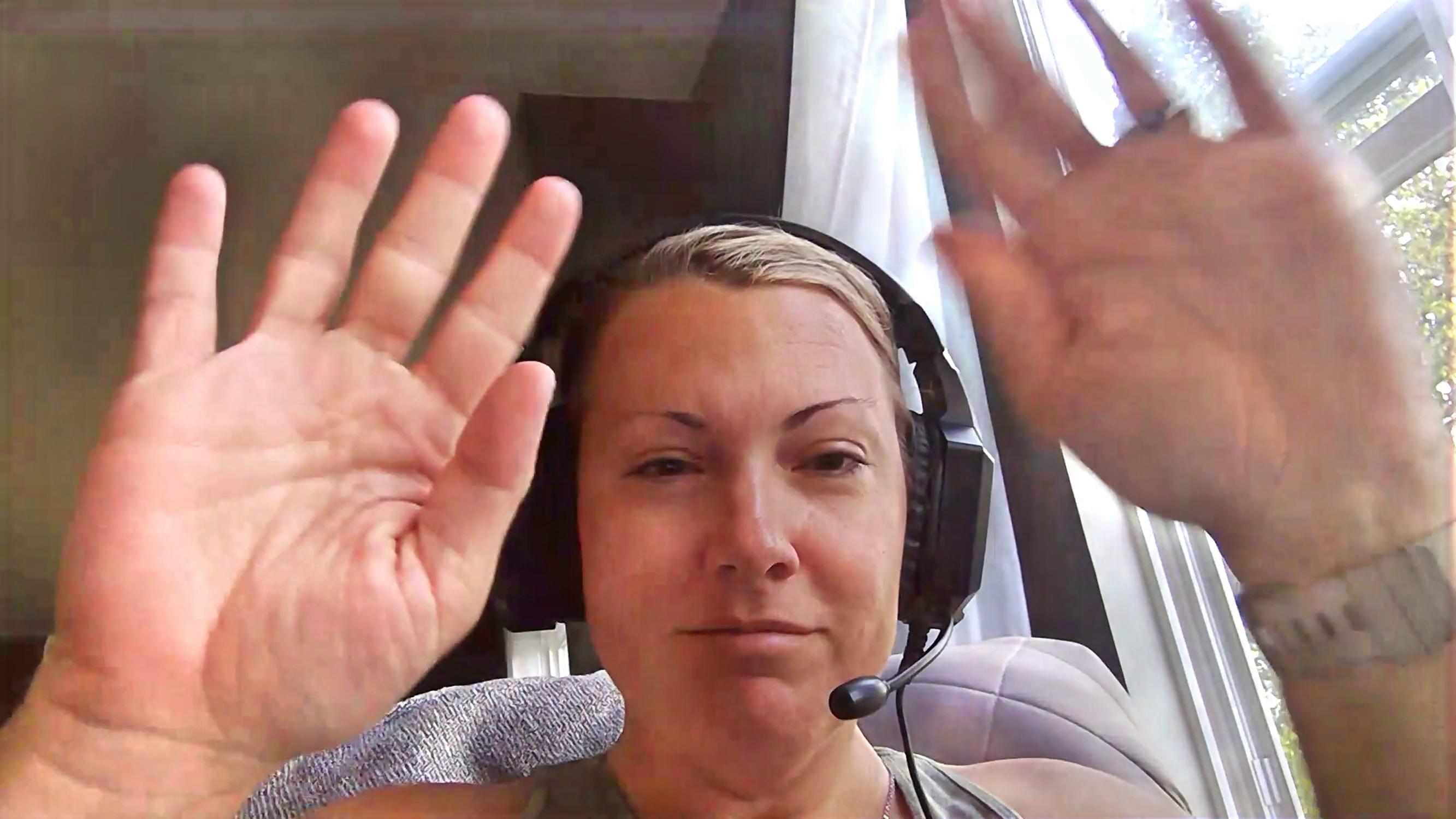} &
        \includegraphics[width=0.30\columnwidth]{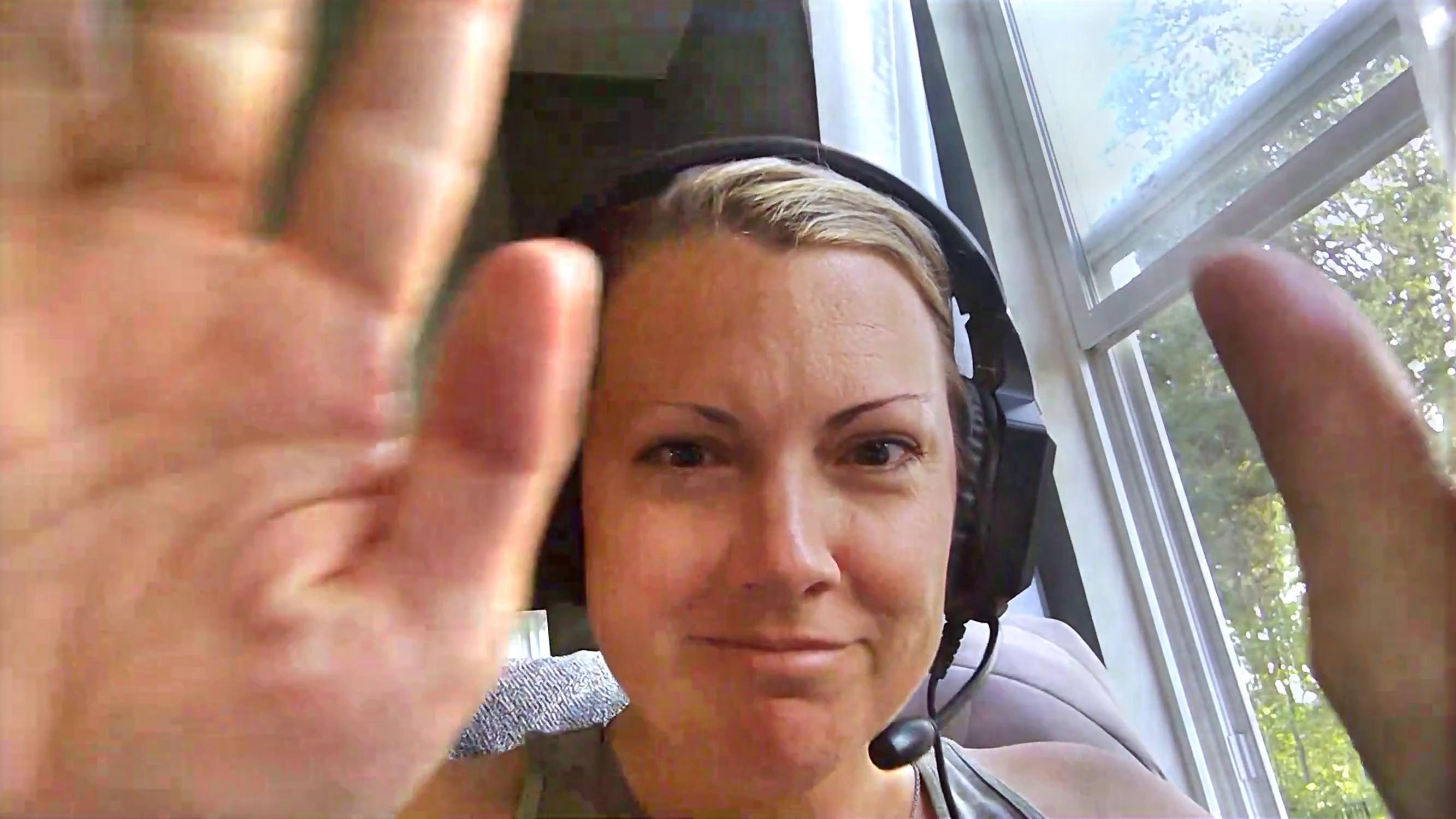} &
        \includegraphics[width=0.30\columnwidth]{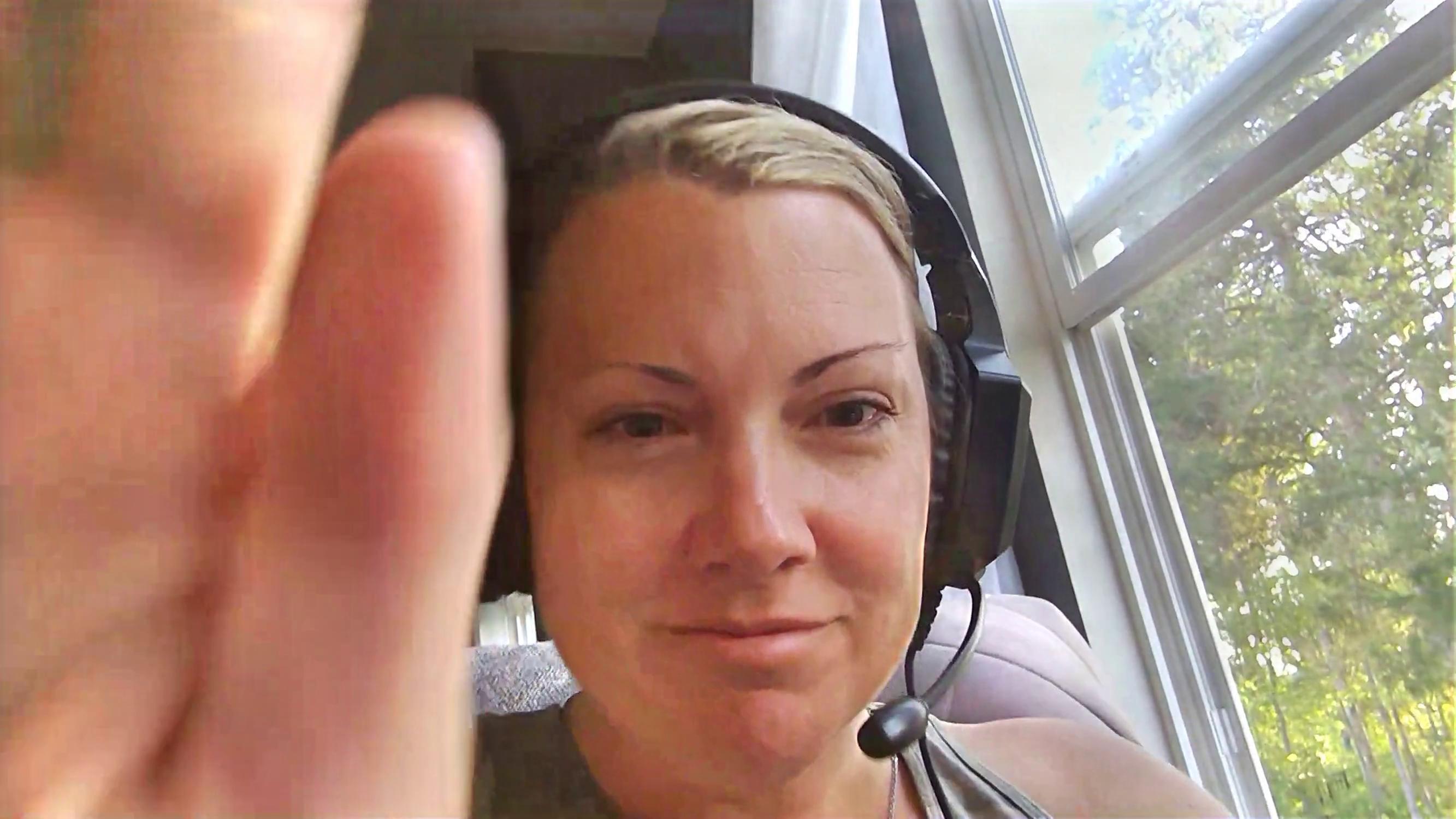} 
    \end{tabular}
    \caption{Comparison of frame enhancement on an unbalanced exposure video from the VCD dataset. From top to bottom: input frames and results of ZeroDCE++, SNRNet, StableLLVE, and our method.}
    \label{fig:video_comparison}
\end{figure}

For real-time video applications, the VisualComparison in Fig.~\ref{fig:video_comparison} (against Zero-DCE++, SNRNet, and StableLLVE) shows that RRNet maintains harmony in skin tone and eliminates overexposed regions that other methods fail to address. Fig.~\ref{fig:processing_time_NIQE} reports per-frame NIQE and end-to-end processing time; since RRNet does not require per-frame lighting estimation, we evaluate three update cadences (1-/3-/10-frame).

Table~\ref{tab:video_temporal} reports the RAFT-based flow-warping error, flicker ratio, and per-frame inference time on VCD clips. Because RRNet represents lighting as a compact parameter vector directly smoothed by the EMA (Eq.~\ref{eq:smoother}) and the renderer is a deterministic per-pixel function of input and depth, no learned dense decoder introduces frame-to-frame stochasticity. At its default 1-frame cadence RRNet already attains flicker $1.35$ and T-Warp $1.36$ — comparable to Zero-DCE++ ($1.46$) and substantially better than FastLLVE ($2.17$) and SNRNet ($1.71$), though still trailing StableLLVE ($1.12$). Lengthening the lighting-update interval from 1 to 10 frames cuts inference from $17.0$ to $2.6$\,ms ($6.5\times$ speedup) \emph{and} further reduces flicker to $1.11$ and T-Warp to $1.17$ — at this 10-frame setting RRNet matches the second-fastest baseline (StableLLVE, $2.5$\,ms) in speed while attaining the best flicker overall (narrowly ahead of StableLLVE's $1.12$) and the second-best T-Warp. The improvement under longer intervals is consistent with the EMA dynamics at $\beta=0.95$: each invocation injects only $5\%$ new observation, so re-estimating less often produces a smoother trajectory of lighting parameters — stability and compute improve together rather than trading off. Together with the per-frame quality results in Table~\ref{tab:comparison} (best 6-set NIQE average), this places RRNet (10-frame) on the Pareto front of speed, temporal stability, and per-frame quality.

\begin{table}[!t]
    \centering
    \caption{Temporal consistency on VCD video clips. \textbf{T-Warp} is the RAFT-based flow-warping error on output frames ($\times 10^{-2}$); \textbf{Flicker} is the ratio of output to input warp error ($1.00$ means no added flicker beyond natural pixel motion). RRNet additionally exposes a configurable lighting-update interval at $\beta=0.95$.}
    \label{tab:video_temporal}
    \footnotesize
    \setlength{\tabcolsep}{4pt}
    \renewcommand{\arraystretch}{1.15}
    \begin{tabular}{lccc}
        \hline
        \textbf{Method} & \textbf{Time (ms) $\downarrow$} & \textbf{T-Warp $\downarrow$} & \textbf{Flicker $\downarrow$} \\
        \hline
        Input (reference) & --   & 1.05 & 1.00 \\
        \hline
        Zero-DCE++   & 1.2  & 1.37 & 1.46 \\
        SNRNet       & 29.0 & 1.58 & 1.71 \\
        FastLLVE     & 7.4  & 1.77 & 2.17 \\
        StableLLVE   & 2.5  & \textbf{1.11} & \underline{1.12} \\
        \hline
        RRNet (1-frame, default) & 17.0 & 1.36 & 1.35 \\
        RRNet (3-frame)          & 6.3  & 1.24 & 1.19 \\
        RRNet (10-frame)         & 2.6  & \underline{1.17} & \textbf{1.11} \\
        \hline
    \end{tabular}
    \vspace{2pt}
\end{table}

\subsection{Glare Removal}
\label{subsec:glare_removal}

\begin{figure}[!t]
    \centering
    \setlength{\tabcolsep}{1pt}
    \footnotesize
    \begin{tabular}{cccc}
        \includegraphics[width=0.22\columnwidth]{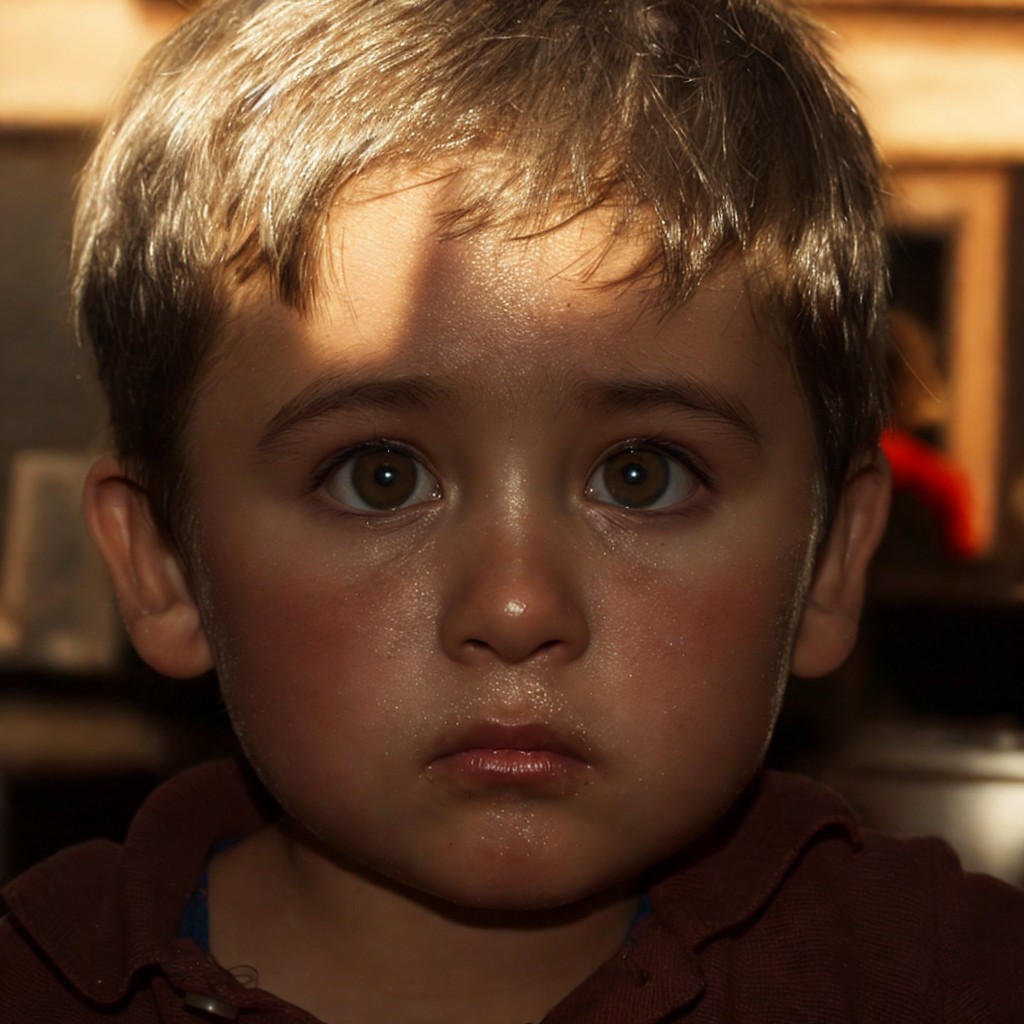} &
        \includegraphics[width=0.22\columnwidth]{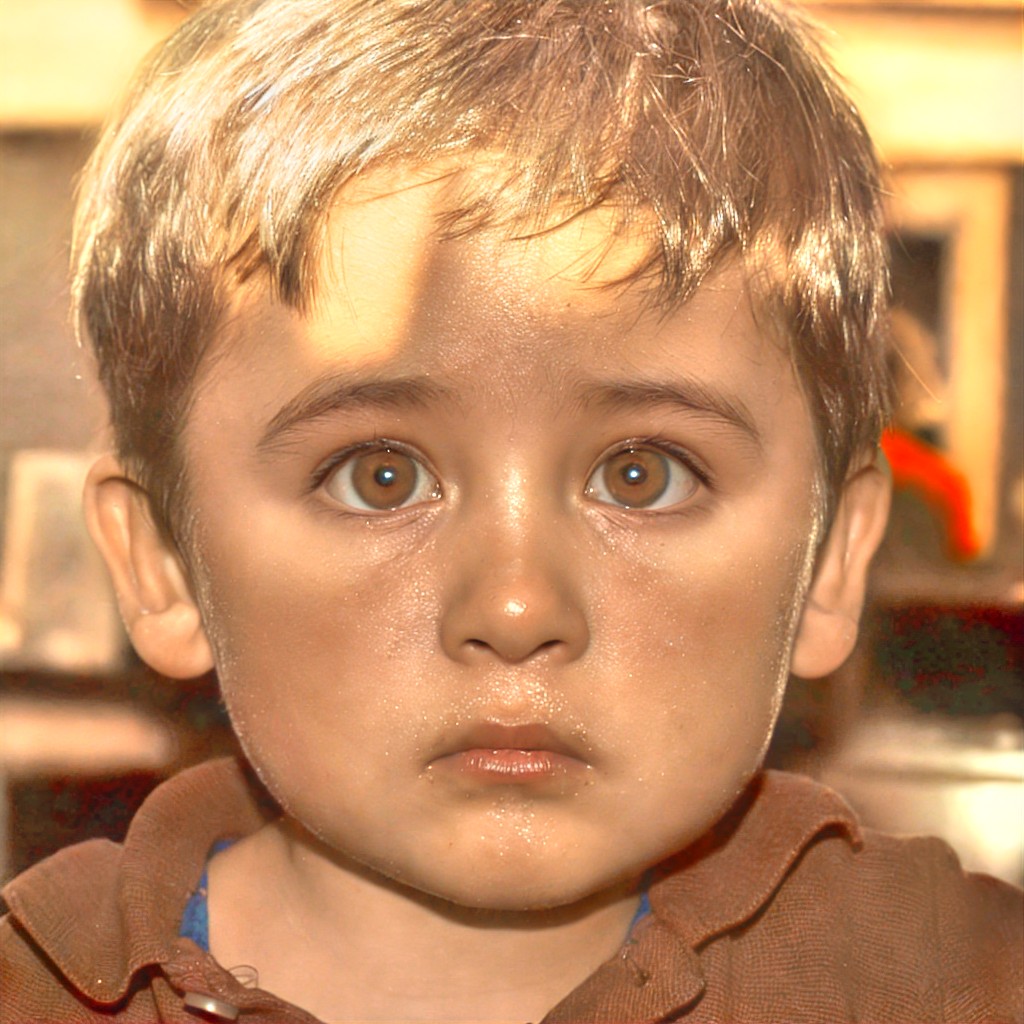} &
        \includegraphics[width=0.22\columnwidth]{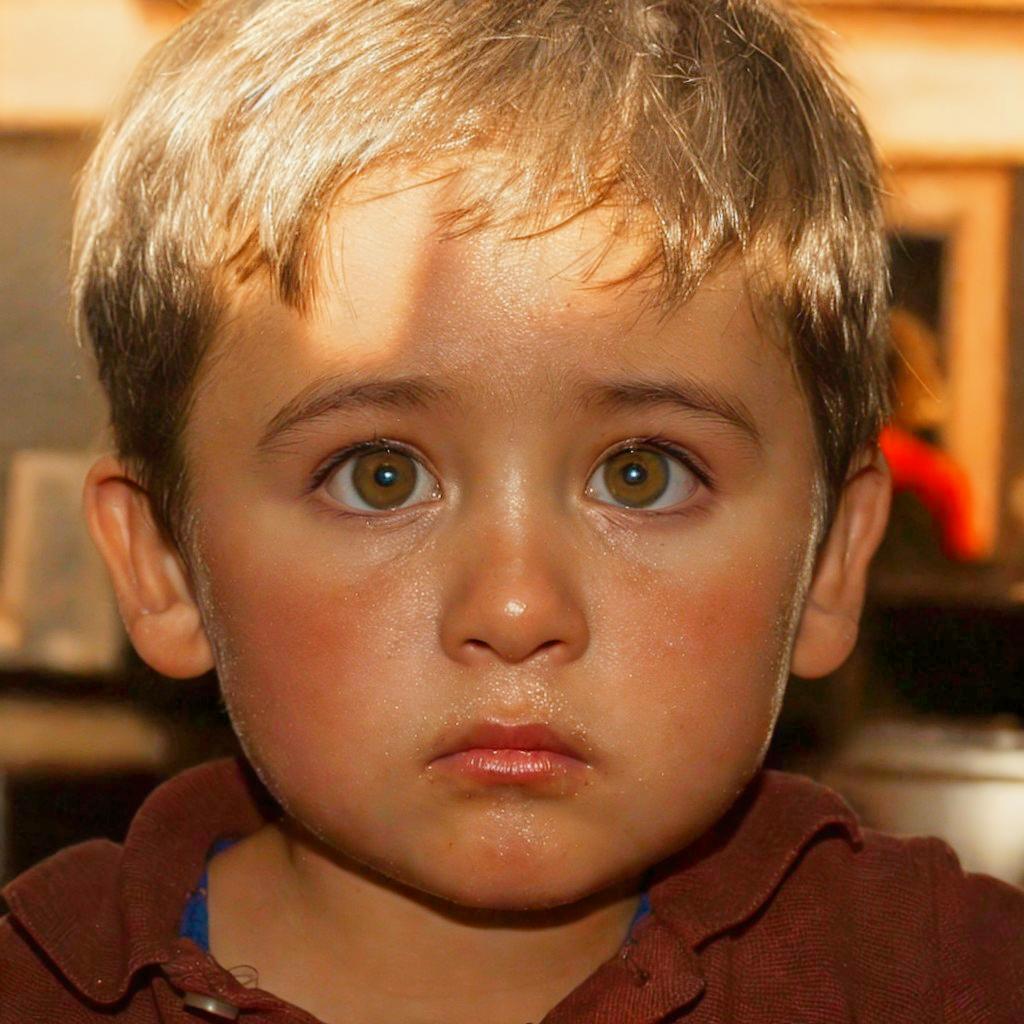} &
        \includegraphics[width=0.22\columnwidth]{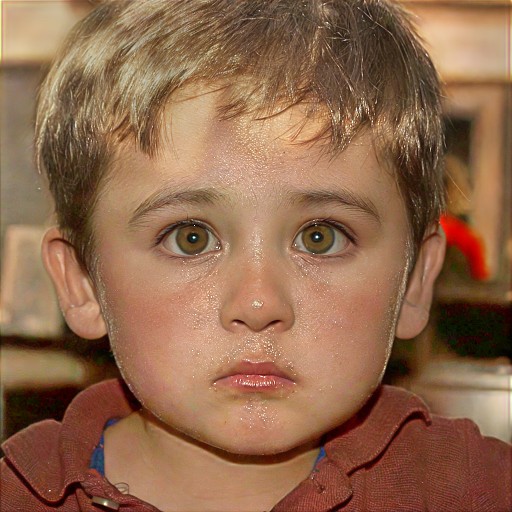} \\
        Input & EnlightenGAN & CIDNet & \textbf{RRNet} \\
    \end{tabular}
    \caption{Qualitative glare removal: with AGM enabled, RRNet's renderer operates on the estimated albedo and locally suppresses strong directional sources while preserving facial structure that pixel-wise baselines either dim or amplify.}
    \label{fig:deglare}
\end{figure}

RRNet generalizes beyond low-light enhancement. With AGM enabled, the renderer operates on the estimated albedo instead of the raw input, so localized highlights are attenuated rather than enhanced (Fig.~\ref{fig:deglare}), with no architectural change beyond toggling AGM on.

\section{Conclusion}
\label{sec:conclusion}

We propose RRNet, a real-time rendering--relighting network that enhances images and videos under unbalanced lighting by regressing virtual lighting parameters and rendering with depth-aware shading. Its lightweight, decoder-free design enables localized, identity-preserving relighting trained from unaligned synthetic data, achieving the best 6-set NIQE average and stable per-set quality at $17$\,ms per frame (down to $2.6$\,ms with longer update intervals) — a practical and scalable solution for real-time video enhancement.

\bibliographystyle{IEEEbib}
\bibliography{main}

\end{document}